# ALMA MATER STUDIORUM
# UNIVERSITÀ DI BOLOGNA

______________________________________________

### DEPARTMENT OF COMPUTER SCIENCE AND ENGINEERING

ARTIFICIAL INTELLIGENCE

### MASTER THESIS

in

Mathematical Methods for Artificial Intelligence

# FRUGALITY IN SECOND-ORDER OPTIMIZATION: FLOATING-POINT APPROXIMATIONS FOR NEWTON'S METHOD

CANDIDATE

Giuseppe Carrino

SUPERVISOR

Prof. Elena Loli Piccolomini

CO-SUPERVISORS

Prof. Elisa Riccietti, ENS Lyon,

Prof. Theo Mary, Sorbonne Univ.

*A Virginia*

# Contents

# Chapter 1

# Introduction

Supervised *Machine Learning* is an Artificial Intelligence field, whose methods consist of the training of predictive models using datasets of observations. The models themselves are parametric functions that take some features as input, and whose output is expected to be close to the observed targets; to achieve this, a **loss** - or *cost* - function is defined, and the training task consists of minimizing it by tuning the model's parameters, which amount to the solution of an **optimization** problem.

Figure 1.1: Weight parameters increase in the last century

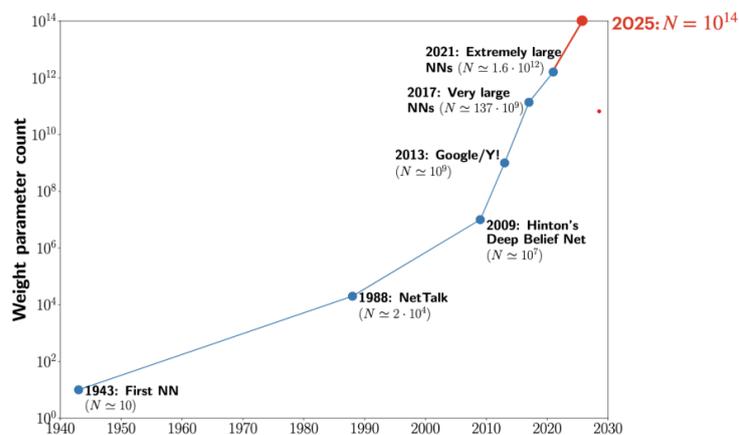

An emerging problem in this field is the *computational cost* of minimization: in the last century, the number of parameters to estimate during the training process increased almost exponentially, as a consequence of big architectures arising, such as *Transformers* [48]. As visible in *Figure* 1.2, taken from [49], also datasets' size is substantially increasing in the last 50 years, making training even harder from both a *time* and *space* consumption perspective.

Figure 1.2: Datasets' size increase over time [49]

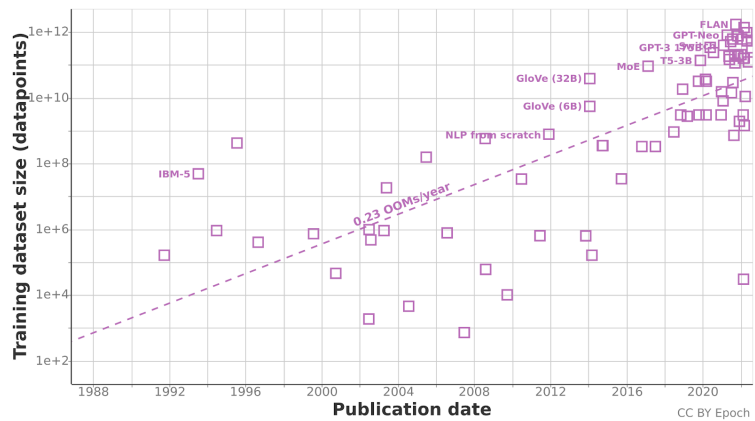

Training modern machine learning models often involves solving large-scale, ill-conditioned optimization problems where first-order methods, such as stochastic or batch gradient descent, can suffer from slow convergence and high sensitivity to the choice of learning rate [23]. In such settings, more advanced optimization strategies are needed to accelerate convergence and improve robustness. Second-order methods, such as Newton's method, address these challenges by incorporating curvature information through the Hessian, enabling them to take more informed steps that automatically account for problem geometry. While gradient descent typically exhibits linear convergence, Newton's method achieves quadratic convergence in a neighborhood of the optimum [36, 4], often resulting in far fewer iterations and reduced sensitivity to hyperparameter tuning. In the context of machine learning, these benefits have been demonstrated in various works: Martens [34] showed that Hessian-based approaches can accelerate deep network training; Bottou et al. [3] highlight that second-order methods improve stability in large-scale learning; and Byrd et al. [5] demonstrate that Hessian subsampling can make them scalable to massive datasets. The method has been used to train Machine Learning models in the past, especially in its *Inexact* [37] and *Quasi-Newton* [15, 42, 29, 35] variants, showing how second-order information can help achieving impressive results especially in the Physics-Informed Neural Networks settings [30, 18]. In particular, different works are training neural networks using Gauss-Newton [40, 2], also in a stochastic setting [13]. Overall, the exploitation of second-order information can be essential for achieving both reliable and efficient training in challenging optimization landscapes. However, Newton's method comes with a higher cost than gradient descent, mainly due to the computation of second-order derivatives and linear system solution. In recent years, the use of low-precision arithmetic has emerged as a successful strategy to reduce optimizers' computational costs, driven by the development of specialized hardware accelerators for machine learning, such as Google's TPUs [26] and NVIDIA's tensor cores [8]. These recent advances have, in fact, made it possible to efficiently exploit low precisions, fostering the rise of quantization as a means of reducing resource requirements. However, the blind use of low precision can lead to significant performance losses. It is therefore essential to understand where and how to use low precision to reduce the computational cost as much as possible while avoiding numerical issues and preserving model accuracy. In the field of optimization, little attention has been paid to **mixed-precision** strategies, with only a few preliminary studies [9, 17]. However, not all steps of an optimization process are equally sensitive to numerical perturbations, and using the same precision for all of them leads to choosing an unnecessarily high precision dictated by the most sensitive steps. For example, in a neural network, not all weights and layers are equally sensitive to numerical perturbations; similarly, in an iterative process, not all steps are equally sensitive either. This has led, in particular, to the development of variable-precision quantization methods for neural networks [12, 11, 16, 31, 51, 52, 50, 47], as well as mixed-precision linear algebra methods—see [21] for a complete review. By working directly on floating-point precision, it is possible to significantly reduce the energy consumption of training [32, 25], other than allowing the training in-place on embedded devices [14, 22], and, combined with second-order optimizers, can do so without sacrificing the model's accuracy and training convergence speed. The main challenge lies in determining the appropriate precision for each step of the iterative process.

A rigorous error analysis for Newton's method in finite arithmetic exists for the variant developed to solve nonlinear systems [28, 27, 45], i.e., to find a zero $x \in \mathbb{R}^n$ of a vector function $F : \mathbb{R}^n \mapsto \mathbb{R}^m$:

$$F(x) = 0 \tag{1.1}$$

This analysis can guide the proposal of a mixed-precision variant of the method. However, this variant of Newton's method is not suitable for tackling machine learning problems, which are rather of the form

$$\min_x f(x) \tag{1.2}$$

for a scalar function $f : \mathbb{R}^n \mapsto \mathbb{R}$. Another variant of Newton's method has been developed for this type of problem, known as Newton's method for optimization [36]. Based on this definition of Newton's method for minimization, the contribution of this work is twofold:

- We developed a **mixed-precision Newton's algorithm for minimization**, reducing the cost of the minimization by using significantly fewer bits per parameter at training time;

- We proposed a Quasi-Newton method that generalizes the Gauss-Newton algorithm, namely **GN_k**, avoiding the computation of less useful second-order derivatives.

This will be done from three perspectives:

1. **Analytical**, by showing convergence proofs for our minimization methods;

2. **Algorithmic**, by providing practical implementations of the optimizers;

3. **Applicative**, by using these algorithms on some real-world classification tasks.

# Chapter 2

# Background

To better understand what follows in this work, and also to provide a starting formalization, this chapter offers a brief technical introduction to the basic concepts of mixed-precision minimization using Newton's method, in addition to introducing its applicative context for regression problems.

## 2.1 Newton's method for optimization

Newton's method is a second-order minimization algorithm. It exhibits quadratic convergence near the solution, thus needing significantly fewer iterations than first-order methods, such as Gradient Descent. A formalization and convergence proof for the method can be found in [36, Thm. 3.5].

First, we should introduce a general minimization problem: let $f : \mathbb{R}^n \mapsto \mathbb{R}$ be twice continuously differentiable in $\mathbb{R}^n$, our minimization problem is

$$\min_{x \in \mathbb{R}^n} f(x). \tag{2.1}$$

We denote by $g$ the gradient $(\partial f / \partial x_i)$ of $f$ and with $H$ $(\partial^2 f / \partial x_i \partial x_j)$ the Hessian matrix of $f$.

We assume that $H$ is positive definite and Lipschitz continuous with constant $L_H$ in $\mathbb{R}^n$, that is,

$$\|H(w) - H(v)\| \leq L_H \|w - v\| \quad \text{for all} \quad v, w \in \mathbb{R}^n \tag{2.2}$$

where $\|\cdot\|$ denotes any vector norm and the corresponding operator norm.

To solve (2.1), Newton's algorithm leverages an iterative approach, improving the solution $x_i$ at each step. It is defined as

$$\begin{aligned} \text{Solve } H(x_i)d_i &= -g(x_i) \\ x_{i+1} &= x_i + d_i \end{aligned} \tag{2.3}$$

We define $x^*$ as the solution for the minimization problem, and write $g_*$ and $H_*$ to indicate, respectively, the gradient and Hessian computed at $x^*$. We will generically denote $x_i$ as $x$ and $x_{i+1}$ as $\bar{x}$ in theorems. Accordingly, we refer to $g(x_i)$ as $g$ and to $H(x_i)$ as $H$. With $\kappa(\cdot)$ we denote the condition number of a matrix.

It is also useful to define two main approximated variants of Newton's method, namely

- **Quasi-Newton**, which consists of the usage of an approximation of the Hessian matrix, denoted as $B$. Known examples are BFGS [36, Ch. 6.1] and Gauss-Newton [10, Ch. 10.2];

- **Inexact Newton**, whose idea is to solve the step computation linear system using an iterative linear solver - a common choice is *Conjugate Gradient*, hence including an error given by the chosen stopping condition, as follows

$$\begin{aligned} r_i &= H(x_i)d_i + g(x_i), \\ \|r_i\| &\leq \eta \|g(x_i)\|, \end{aligned} \tag{2.4}$$

where $\mathbb{R} \ni \eta \ll 1$ is the *stopping condition* hyper-parameter.

Similarly to Gradient Descent, the method leverages the negative gradient as the descending direction, but it also includes second-order information, which we can think of as the *curvature* of the function, to speed up the convergence: this comes, of course, with a higher computational cost due to

- Hessian evaluations;

- Linear system solving - which can be non-trivial;

- Usage of high numerical precision, to avoid numerical instability.

## 2.2   Floating-point arithmetic

When we talk about *finite precision* in the computer science framework, what we refer to is the representation of real numbers, whose number of digits can be infinitely large, on a finite-memory machine, such as our personal computers. The common representation methodology for real numbers is *floating-point* representation, which can be thought of as a sort of scientific notation, and involves three elements:

- a **significand**,

- a **base**,

- an **exponent**.

For example,

$$2469/200 = 12.345 = \underbrace{12345}_{\text{significand}} \times \underbrace{10}_{\text{base}}{}^{\overbrace{-3}^{\text{exponent}}}. \qquad (2.5)$$

Note that on our usual hardware, the used base is $2$ instead of $10$.

On a finite-memory machine, there are constraints on the actual representable numbers: based on the amount of available bits, we cannot represent numbers smaller than a certain value, which is usually referred to as *machine epsilon*, or **unit roundoff**, and denoted as $u$. The IEEE standardized the computer representation for binary floating-point numbers in IEEE 754 in 1985, and subsequently revised it in 2008. The final classification is summarized in the following table.

| **Number of bits** | **Signif. (t)** | **Exp. range** | $u = 2^{-t}$ |
|---|---|---|---|
| fp128 (quadruple) | 113 | $10^{\pm 4932}$ | $1 \times 10^{-34}$ |
| fp64 (double) | 53 | $10^{\pm 308}$ | $1 \times 10^{-16}$ |
| fp32 (single) | 24 | $10^{\pm 38}$ | $6 \times 10^{-8}$ |
| fp16 | 11 | $10^{\pm 5}$ | $5 \times 10^{-4}$ |
| bfloat16 (half) | 8 | $10^{\pm 38}$ | $4 \times 10^{-3}$ |
| fp8 (e4m3) | 4 | $10^{\pm 2}$ | $6 \times 10^{-2}$ |
| fp8 (e5m2) (quarter) | 3 | $10^{\pm 5}$ | $1 \times 10^{-1}$ |

Table 2.1: Floating point formats and their characteristics.

Here, we can see how, trivially, the more bits we use to represent a number, the bigger the significand can be, and so the smaller the $u$ becomes. Also, more bits allow us to represent more numbers, as we can see from the *exponent range* column. Due to the finite memory, almost[1] every number comes with an intrinsic error, due to a partial loss of its decimal digits. This error intuitively propagates when finitely represented numbers go through arithmetic operations: as extensively shown in [20], arithmetic operations' error follows the equation

$$fl(x \text{ op } y) = (x \text{ op } y)(1 + \delta), \quad |\delta| \leq u, \quad \text{op} = +, -, \times, /, \qquad (2.6)$$

where $fl(\cdot)$ is the function that we use to transform a value represented in exact arithmetic into a floating-point number. For simplicity, from now on, a variable $x$ represented using floating-point methodology will be noted as $\hat{x}$.

This means that, for each operation, we make a relative error $\delta$ which is at most as big as the unit roundoff of the chosen precision, but it also implies that the error propagates the more the operations we do, and so that in a complex algorithm the resulting finite-precision approximation error can be significantly higher than the machine epsilon. This practically impacts the convergence of optimization algorithms, and it's the reason why high precisions are chosen for algorithms such as Newton's method.

---

[1]The so-called *machine numbers* can be represented with no error. An example is the integers;

In a mixed precision context, which is the one we want to work in, we don't focus on *the* precision, but on multiple floating point precisions used for different operations: these will be noted using

- $\pi_X$, as the used precision;

- $u_X$, as the unit roundoff of precision $\pi_X$,

where $u_X \in \mathbb{R}$, $\pi_X \in \{16, 32, 64, 128\}$ and $X \in \{g, w, l\}$. These last pedices refer to different sections of Newton's method in which a specific precision will be used. In particular, $g$ stands for $gradient$, $w$ stands for $working$, and $l$ stands for $linear$. In *Chapter* 3, this division will be clearer and justified by intuitions and error analysis.

## 2.3  Logistic Regression

For the sake of completeness, we also briefly introduce *logistic regression* base intuitions.

In general, we can use optimization algorithms to perform *regressions*, i.e., estimate parameters of some model functions based on a set of observed datapoints $(X_i, y_i) \in \mathbb{R}^m \times \mathbb{R}$. This is done by minimizing an *objective*, or **loss**, function, which represents the distance between the predictions of our model and the *true* target values. Formally, we can rethink our function $f(x)$ as parametrized by parameters $\theta \in \mathbb{R}^d$, and dependent on a model function $F_X : \mathbb{R}^d \mapsto \mathbb{R}^n$, such as

$$f_X(\theta) = \frac{1}{N} \sum_{i=1}^{n} (F_{X_i}(\theta) - y_i)^2 \qquad (2.7)$$

where $X \in \mathbb{R}^{n \times m}$, $y \in \mathbb{R}^n$ are the observations. When minimizing $f_X(\theta)$ with respect to $\theta$ using an algorithm such as Newton's, we are finding parameters for $F_X$ that make the model's output as close as possible to $y$.

Based on what type of observations we have and what predictions we want to output, different models and loss functions are used. In Machine Learning, a simple task is to train a model that predicts a binary target class: for each $X_i$, the model outputs a real value between $0$ and $1$, and thus, using a threshold, a binary output is obtained. An example of this type of problem is represented by the MUSH dataset [46], where each entry - a mushroom - with $m$ features, is labeled as *edible* or *poisonous*. The task here is to train a model that, given these $m$ features, outputs the edibility of the mushroom. Now, the question is: what do we use as a model $F$? In Deep Learning, $F$ is a neural network. For simpler problems, simpler models can be used, and this is the case for *Logistic Regression*, where a sigmoid-based function is employed and has proven to work effectively for a wide range of datasets. Formally,

$$F_X(\theta) = \underbrace{\sigma}_{\text{sigmoid}}(X\theta) = \frac{1}{1 + e^{-X\theta}}, \tag{2.8}$$

which needs the dimensions of $\theta$, i.e., $d$, to be equal to the number of features of input entries, i.e., $m$. For this model function, and the setting of a *Binary Classification task*, the commonly used loss function is the **Binary Cross-Entropy**, or **Negative Log-Likelihood**,

$$\begin{aligned} BCE(y, \hat{y}) &= -\frac{1}{N} \sum_{i=1}^{N} y_i \log(\hat{y}_i) + (1 - y_i) \log(1 - \hat{y}_i), \\ \hat{y} &= F_X(\theta). \end{aligned} \tag{2.9}$$

Therefore, to train a model for a binary classification task, all that is needed is an initial parameter guess and a minimization algorithm to be run on the $BCE$ function, such as Newton's method.

# Chapter 3

# Mixed Precision Newton

The purpose of this chapter is to present analytical and practical results for a mixed-precision approach applied to Newton's minimization method. As explained in the Introduction (see 1), Newton's method is used to find zeros of a function: an error analysis of its usage in a Mixed-Precision context does exist in [21] and [45]. However, the study focuses on the usage of Newton for solving $f(x) = 0$: in this Chapter, we will adapt the related theorems to the minimization context, instead, where we aim to solve $\nabla f(x) = 0$. The main questions are

**Q1.** Which precisions shall we use in each step of Newton's method?

**Q2.** Can we extend mixed-precision results to other Newton's method approximations?

**Q3.** Does mixed-precision Newton work for machine learning?

First, we write a general Newton's method pseudo-code. Since we want our analysis to be general enough, we will work on Quasi-Inexact-Newton, taking into account that linear system solving can introduce an approximation error due to iterative methods, and that the Hessian can be a close approximation.

Hence, we highlighted three main sections of the algorithm:

---

**Algorithm 1** Mixed-Precision Quasi-Inexact-Newton

    **for** $i = 1 : maxit$ or until converged **do**
        Compute $g_i = g(x_i) \leftarrow$ in precision $\pi_g$
        Solve $H(x_i)d_i = -g_i \leftarrow$ in precision $\pi_l$
        $x_{i+1} = x_i + d_i \leftarrow$ in precision $\pi_w$
    **end for**

---

1. the **gradient computation**, in precision $\pi_g$,

2. the **linear system solving**, in precision $\pi_l$,

3. the **step addition**, in precision $\pi_w$, which stands for *working*.

This division, which is derived from the work of [21] on the root-finding Newton's method, comes with some intuitions:

- The gradient computation is the most important step, since it can be very error-prone due to its arithmetic complexity, or its approximation, and a wrong direction for the optimization can also lead to non-minimization of the function;

- Linear system solving error should be of a lower magnitude than the one coming from non-linear steps;

- The relative error on $x_{i+1}$ will *never* be lower than $u_w$ (see 2.6).

To now answer precisely **Q1** and **Q2**, we will proceed with an error analysis of our algorithm, adapted from [45] to the minimization context. Then, experiments with a logistic regression will be performed to deal with **Q3**.

## 3.1 Error Analysis

First, we can write Newton's iterative step as a single equation, separating variables from their relative errors, which arise from finite precision representation or Quasi-Inexact Newton approximations (see 2.1). We see that our next solution guess $x_{i+1}$ can be defined as follows

$$\hat{x}_{i+1} = \hat{x}_i - (H(\hat{x}_i) + E_i^H)^{-1}(g(\hat{x}_i) + E^g) + E^+, \qquad (3.1)$$

where $\hat{x}$ is $x$ including errors coming from rounding (as in *Sec.* 2.2) and other errors, explained below.

- $E_i^g$ is the error made in computing $g(\hat{x}_i)$ and we assume that there is a function $\psi$ depending of $g$, $\hat{x}_i$, $u$ and $u_g$ such that

$$\|E_i^g\| \leq u\|g(\hat{x}_i)\| + \psi(g, \hat{x}_i, u, u_g), \qquad (3.2)$$

This error is so the same as $e$ defined in [45],

- $E_i^H$ combines the error incurred in forming $H(\hat{x}_i)$ with the backward error for solving the linear system for $d_i$, other than considering the error of approximating the Hessian using a Quasi-Newton approach. We assume that there is a function $\phi$ that reflects both the (in)stability of the linear system solver and the error made when approximating or forming $H(\hat{x}_i)$, such that

$$\|E_i^H\| \leq \phi(g, \hat{x}_i, n, u_l, u)\|H(\hat{x}_i)\|, \qquad (3.3)$$

In practice, we certainly have $\phi(g, \hat{x}_i, n, u_l, u) \geq u_l$, which means that this error is at least the one made by using finite precision,

- $E^+$ is the rounding error made when adding the correction $\hat{d}_i$ to $\hat{x}_i$, so

$$\|E_i^+\| \leq u(\|\hat{x}_i\| + \|\hat{d}_i\|). \qquad (3.4)$$

Starting from this equation, we aim to derive the convergence inequality of the approximated Newton method, to identify which error terms most significantly impact the relative error between $x_i$ and the solution $x^*$.

**Theorem 3.1.1.** *Assume that there is a $x^*$ such that $g(x^*) = 0$, $H_* = H(x^*)$ is nonsingular, and*

$$u_l \kappa(H(\hat{x}_i)) \leq \phi(g, \hat{x}_i, n, u_l, u) \kappa(H(\hat{x}_i)) \leq \frac{1}{8} \quad \text{for all } x_i \text{ and } x^*. \quad (3.5)$$

*Then, for all $x$ such that*

$$L_H \|H_*^{-1}\| \|x - x^*\| \leq \frac{1}{8}, \quad (3.6)$$

*$\bar{x}$ is well defined and Newton's method in floating point arithmetic generates a sequence $\{\bar{x}\}$ whose normwise relative error decreases until the first $i$ for which*

$$\frac{\|\bar{x} - x^*\|}{\|x^*\|} \approx lim_{acc} \triangleq \frac{\|H_*^{-1}\|}{\|x^*\|} \psi(g, x^*, u, u_g) + u. \quad (3.7)$$

Proof of the theorem can be found in *Appendix* A.

We can also replicate this bound derivation for the gradient norm, as we usually focus on reaching a *good enough* local minimum point, rather than the exact solution $x^*$.

First, we can already show by intuition a non-sharp bound on the gradient. As before, we write $g = g(x)$ and $H = H(x)$. Reminding that $\hat{x}^* = fl(x^*) = x^* + \Delta x^*$ with $\|\Delta x^*\| \leq u\|x^*\|$ (see 2.6), then *Lemma* A.0.1 gives

$$g(\hat{x}^*) = g(x^* + \Delta x^*) = H(x^*)\Delta x^* + \epsilon, \text{ where } \|\epsilon\| \leq \frac{L_H}{2}\|\hat{x}^* - x^*\|^2,$$

thus

$$\|g(\hat{x}^*)\| \leq u\|H(x^*)\|\|x^*\| + \frac{L_H}{2}u^2\|x^*\|^2$$

is a non-sharp on the norm of the gradient.

**Theorem 3.1.2.** *Assume that there is a $x^*$ such that it solves the problem*

$\min_{x \in \mathbb{R}^n} F(x)$, so that $g(x^*) = g_* = 0$ and $H(x^*) = H_*$ is nonsingular, and

$$u_l \kappa(H(\hat{x}_i)) \leq \phi(g, \hat{x}_i, n, u_l, u) \kappa(H(\hat{x}_i)) < \frac{1}{8} \quad \text{for all } x_i \text{ and } x^*. \quad (3.8)$$

*Then, if, for all $x$, the limiting accuracy $lim_{acc} \approx \|H_*^{-1}\| \psi(g, x^*, u, u_g) + u\|x^*\|$ satisfies*

$$L_H lim_{acc} \|H_*^{-1}\| < 1/8, \quad (3.9)$$

*and*

$$L_H \|H_*^{-1}\| \|x - x^*\| < \frac{1}{8}, \quad (3.10)$$

*the sequence $\{g(\hat{x}_i)\}$ of residual norms generated by Newton's method in floating point arithmetic decreases until*

$$\|g(\hat{x}_{i+1})\| \approx lim_g \triangleq \psi(g, \hat{x}_i, u, u_g) + u\|H(\hat{x}_i)\| \|\hat{x}_i\| \quad (3.11)$$

Proof of the theorem can be found in *Appendix* B.

Both theorems show us that the error that mostly impacts the quality of the achieved solution is the one coming from $\psi$, i.e., the error on the gradient evaluation, as it was suspected at the beginning of *Chapter* 3. Also, we find that the machine precision of $\pi_w$ is a lower bound for both gradient norm and relative error, as expected. Given these results, we conclude that

- $\pi_g$ must be the highest possible precision,

- $\pi_l$ can be low, since it does not impact how good the achieved solution is,

- $\pi_w$ can be lower than $\pi_g$, but not too low, as it is a lower bound on the final error.

Hence, we can write

$$u_g \leq u_w \leq u_l, \quad (3.12)$$

which means, recalling *Sec.* 2.2, that the higher the precision $\pi_X$, the smaller the machine epsilon $u_X$. Since machine epsilon $u_X$ is also an upper-bound to the error (*Eq.* (2.6)), a higher precision means a lower error. We can now rewrite Mixed-Precision Newton's algorithm as follows,

---
**Algorithm 2** Mixed-Precision Quasi-Inexact-Newton
---
    **for** $i = 1 : maxit$ or until converged **do**
        Compute $g_i = g(x_i) \leftarrow$ in high-precision $\pi_g$
        Solve $H(x_i)d_i = -g_i \leftarrow$ in low-precision $\pi_l$
        $x_{i+1} = x_i + d_i \leftarrow$ in working-precision $\pi_w$
    **end for**
---

This version of the algorithm is what will be used from now on, selecting different precisions and functions based on the experiments. In the next subsection, we will focus on the *Inexact Newton* variant, adapting our convergence theorem to its framework, finally deriving bounds on the iterative solvers' stopping conditions, as defined in [36, Ch. 7.1].

### 3.1.1 Inexact Newton

Newton's method requires the solution of a linear system at each iteration to compute the step, cf 2.3. As highly dimensional linear systems are usually unfeasible to solve with direct methods, and iterative solvers can also help convergence in case of data sparsity, such linear systems are often solved with iterative methods such as conjugate gradients (CG). We will thus present a small enanchement of the analysis of *Section* 3.1 explictly defining the error derived by linear system solving, considering the whole method as a case of *Inexact Newton* - simlarly to what is done for root-finding problem by [28]: this approach will allow us to derive bounds on the iterative solving error only, which are a practical tool when choosing the stopping condition of the linear solver. Additionally, the equivalence between the Inexact Newton model, as derived in [36], and our backward error model on $H$ is provided for completeness in *Appendix* C.

We begin by writing down the Newton step linear system for minimization:

note that we formalize the error in the system solving as a backward error on $H$ (see 3.3). We will consider the *residual* $r$ definition as presented in the Inexact Newton model ([36, Ch. 7]), as follows:

$$\|r\| = \|H\hat{d} + g\| \leq \eta\|g\|. \tag{3.13}$$

Also, as done before, we avoid the explicit writing of $\hat{x}$, thus $H = H(\hat{x})$ and $g = g(\hat{x})$. Hence, we denote $\phi(g, \hat{x}, n, u_l, u)$ simply as $\phi$.

$$\begin{aligned} H\hat{d} + g &= r \\ (H + E^H)\hat{d} = -g &\implies H\hat{d} + E^H\hat{d} = -g. \end{aligned} \tag{3.14}$$

By subtracting the first equation from the second, we have

$$E^H\hat{d} = -r \implies \|E^H\hat{d}\| = \|r\|. \tag{3.15}$$

We now want to relate the error on $H$ to the norm of the gradient. To do so, we use the previous inequalities as follows,

$$\begin{aligned} \|E^H\hat{d}\| &\leq \|E^H\|\|\hat{d}\| \leq \phi\|H\|\|\hat{d}\| = \phi\|H\|\|H^{-1}H\hat{d}\| \\ &\leq \phi\|H\|\|H^{-1}\|\|H\hat{d}\| \\ &\stackrel{\text{using 3.14}}{\leq} \phi\kappa(H)\|-g - E^H\hat{d}\| \\ &\leq \phi\kappa(H)(\|g\| + \|E^H\hat{d}\|). \end{aligned}$$

Starting from this derivation, we can group by $E^H\hat{d}$, as follows,

$$\begin{aligned} \|E^H\hat{d}\| - \phi\kappa(H)\|E^H\hat{d}\| &\leq \phi\kappa(H)\|g\| \implies \\ (1 - \phi\kappa(H))\|E^H\hat{d}\| &\leq \phi\kappa(H)\|g\| \implies \\ \|E^H\hat{d}\| &\leq \frac{\phi\kappa(H)}{(1 - \phi\kappa(H))}\|g\|, \end{aligned}$$

We can see, pointing out that $\|r\| = \|E^H \hat{d}\|$ (see *Eq.* 3.15), and defining,

$$\eta = \frac{\phi \kappa(H)}{(1 - \phi \kappa(H))}, \tag{3.16}$$

that *Inexact Newton* model generalizes our bounding on $\|E^H\|$. Practically speaking, an interesting consequence of this result is that we can relate the upper bounds assumed in *Theorem* 3.1.1 to the stopping condition $\eta$, to bound it too.

$$\begin{aligned}
\eta &= \frac{\phi \kappa(H)}{1 - \phi \kappa(H)} \implies \\
\eta(1 - \phi \kappa(H)) &= \phi \kappa(H) \implies \\
\eta - \eta \phi \kappa(H) &= \phi \kappa(H) \implies \\
\eta &= (\eta + 1)\phi \kappa(H) \implies \\
\phi \kappa(H) &= \frac{\eta}{\eta + 1}.
\end{aligned} \tag{3.17}$$

Hence, using 3.5 and 3.17,

$$\begin{aligned}
\phi \kappa(H) = \frac{\eta}{\eta + 1} &\leq \frac{1}{8} \implies \\
\eta &\leq \frac{1}{7},
\end{aligned} \tag{3.18}$$

which means that, by choosing $\eta \leq 1/7$, we can ensure that the backward error $E^H$ satisfies the conditions required for our theorems.

### 3.1.2  Quasi-Newton

Together with Inexact Newton, there exists a subset of methods approximating Newton called **Quasi-Newton** methods. The idea behind these methods is the use of something different, but close, to the true Hessian matrix of the function being minimized. Many examples exist in literature and are widely used, such as BFGS [44], which is also implemented as a PyTorch optimizer[1]. In this work, we will focus on the Gauss-Newton variant, as it is easily implementable

---
[1] https://docs.pytorch.org/docs/stable/generated/torch.optim.LBFGS.html

and it fits our experiments done on Least Squares problems.

Briefly, the Gauss-Newton method can be applied to Least-Squares problems, working with the inner function, i.e., the *residual*: exploiting problem structure, the method builds a Hessian approximation using gradient information only. In other words, the method leverages the closeness to a solution to avoid computing the second-order differentiation of the residual function, thereby saving both space and time complexity. It has been proven to work in a small residual regime.

From Dennis-Schnabel book [10], a formulation of Non-Linear Least-Squares is

$$\min_{x \in \mathbb{R}^m} f(x) = \frac{1}{2} R(x)^T R(x) \tag{3.19}$$

where $R : \mathbb{R}^m \mapsto \mathbb{R}^n$ is the *residual function*, $r_i(x)$ is the $i$-th component of it, and it is non-linear in $x$. The first derivative of $R$ is nothing but the Jacobian matrix $J(x) \in \mathbb{R}^{n \times m}$, where $J(x)_{ij} = \partial r_i(x)/\partial x_j$. It can be shown that the first derivative of $f$ is

$$\nabla f(x) = J(x)^T R(x), \tag{3.20}$$

and the second derivative is

$$\nabla^2 f(x) = J(x)^T J(x) + S(x), \tag{3.21}$$

where

$$S(x) = \sum_{i=1}^{m} r_i(x) \cdot \nabla^2 r_i(x), \tag{3.22}$$

i.e., $S(x)$ is the term that contains second-order information on the residual. Given this formulation of the minimization problem, we can use Newton's method 3.23 to solve it. The Newton update step would be

$$x_{i+1} = x_i - (J(x_i)^T J(x_i) + S(x_i))^{-1} J(x_i)^T R(x_i). \tag{3.23}$$

Gauss-Newton relies on the definition of the affine model of $R(x)$ around

a point $x_i$:

$$M_i(x) = R(x_i) + J(x_i)(x - x_i), \qquad (3.24)$$

where $M_i : \mathbb{R}^m \mapsto \mathbb{R}^n$, and $n > m$. We cannot expect to analitically find a $x^*$ such that $M_i(x^*) = 0$, but we can decide to minimize $M_i$ to find the next approximated solution $\bar{x}$, as follows,

$$\min_{x \in \mathbb{R}^m} \frac{1}{2} \|M_i(x)\|_2^2 \triangleq \hat{m}_i(x). \qquad (3.25)$$

It is proven in [10, Ch. 3.6] that the solution to this minimization problem is

$$\bar{x} = x - (J(x)^T J(x))^{-1} J(x)^T R(x), \qquad (3.26)$$

which is very similar to the Newton step defined in 3.23, lacking the term $S(x_i)$, which contains the second-order information on the residual function. The lack of this term saves memory and time complexity, aside from avoiding the need for second-order derivatives of the residual function, a characteristic of Quasi-Newton methods.

Since the Newton method is proven to have quadratic convergence, and the Gauss-Newton step differs from it only by the addition of $S(x)$, it is easily deducible that if this term equals zero, quadratic convergence can also be proven for the Gauss-Newton method. This occurs in the case of zero-residual problems, i.e., when the starting point $x_0$ is close to the solution $x^*$, and when $R$ is linear. On the other hand, if $S(x)$ is big w.r.t $J(x)^T J(x)$, the Gauss-Newton method may not be locally convergent at all. These results are formalized in the convergence theorem of Gauss-Newton (see [10, Th. 10.2.1]). The proof shows a super-linear convergence of the method when

$$\|(J(x) - J(x^*))^T R(x^*)\| \leq \sigma \|x - x^*\|, \qquad (3.27)$$

where $\sigma$ is a free parameter lower then $\lambda$, i.e., the minimum eigenvalue of $J(x^*)^T J(x)$. However, it is reported in Dennis-Schnabel's book that this term is approximately equal to $S(x^*)(x - x^*)$. Since we aim to upper-bound the discarded term $S$, we will prove this similarity in *Appendix* D.

We now would like to apply our Mixed-Precision setting to Gauss-Newton. First, it is easily provable that the standard assumptions on the convergence of Gauss-Newton implicitly imply, in the majority of cases, also the assumptions made in our *Theorem* 3.1.1: this is again shown in *Appendix* D. This is a significant result, as it means that for almost all functions for which Gauss-Newton exhibits super-linear convergence, we can utilize our Mixed-Precision approach to compute the attainable relative error and gradient norm a priori, without requiring any additional assumptions. Given our results and the generality of our mixed precision framework, we can provide an algorithm for Gauss-Newton that respects the predicted bounds and convergence rate in *Theorems* 3.1.1 and 3.1.2, as follows,

---
**Algorithm 3** Mixed-Precision Gauss-Newton

    **for** $i = 1 : maxit$ or until converged **do**
        Compute $g_i = J(x_i)^T R(x_i) \leftarrow$ in high-precision $\pi_g$
        Solve $J(x_i)^T J(x_i) d_i = -g_i \leftarrow$ in low-precision $\pi_l$
        $x_{i+1} = x_i + d_i \leftarrow$ in working-precision $\pi_w$
    **end for**

---

which follows the same structure of 2, but computes an approximation of the Hessian.

### 3.1.3 Newton's approximation wrap-up

Wrapping up this chapter of analytical analyses, we can answer questions **Q1** and **Q2**:

**A1.** Through error analysis, we described an algorithm that leverages high precision for high importance steps - i.e,. gradient computation - and low precision for less critical ones - i.e., linear system solving, also giving upper bounds to approximation errors and allowing the computation *a priori* of attainable relative error and gradient norm;

**A2** .Our model already includes in its error definition error possibly coming from Hessian formation (*Quasi-Newton*) and iterative linear system solving (*Inexact Newton*): we also derived practically usable bounds for the solver's stopping condition and show that Gauss-Newton in Mixed-Precision does not require additional assumptions than exact arithmetic ones, to ensure its convergence within the predictable bounds.

## 3.2 Practical Implementation

To assess the soundness of the results obtained by applying Mixed Precision to Newton's method, a practical algorithm has been implemented in Python, leveraging the NumPy library[2]. The implementation did not require essential changes to what we described with pseudo-code in the previous section. To change the precision of real values, NumPy offers a set of functions and floating-point precisions. It must be noted, however, that precision fp128 differs from the IEEE standard: in fact, not all the hardware can represent numbers in *quadruple* precision, thus when we refer to precision 128, we are referring to a sort of *long double* precision, which uses 80 bits instead of 64. To perform tests, abstract classes for implementing Least Squares functions, along with their first and second derivatives, have been provided.[3]

The first experiments will focus on simple Least Squares regression problems, minimizing the function $f_X(\theta)$ as defined in 2.7 with respect to $\theta$, using Mixed-Precision Newton. First, for each experiment, we will define a model function $F_X(\theta)$ and a desired solution $\theta^*$. Then $y$ is computed as $F_X(\theta^*) +$ noise, where the noise is sampled from a uniform distribution. The noise addition is made to make the problem more challenging and also to ensure a non-zero minimum, which would be less desirable in a finite precision context. However, to assess solely the error coming from the mixed-precision approach, a first run of classical Newton's method, using precision np.float128, is done for $\sim 500$ iterations: the solution reached is the one set as $\theta^*$ for the following experiments. In this way, we will have a noisy and more realistic problem; however, when we check the relative error of our achieved solutions, we will do so taking into account the mixed-precision approximations only. After having the problem set up, some starting points $\theta_0$ are chosen not far from the solution. For each experiment, we will first ensure that the bounds defined in *Theorem* 3.1.1 are respected. Then, the predicted $lim_{acc}$ (3.7) and $lim_g$ (3.11) are computed to compare them to the results achieved by the minimizer. To do so, the computation of term $\psi$ - i.e., the error on the gradient computation - is needed. Theoretically speaking, error analysis should be performed on the function gradient to have an analytical formula that describes the error propagation for the specific function. In practice, this method would not be feasible with complex model functions. Hence, the error is computed numerically, in a more approximate, but more easily replicable, way, as the absolute distance between the gradient calculated in the highest possible precision (np.float128) and the used precision, as follows:

$$\psi = \|\text{np.float128}(g_X(\theta^*)) - \pi_g(g_X(\theta^*))\|. \tag{3.28}$$

---

[2]https://numpy.org/
[3]The whole repository, along with the results charts, is available at https://github.com/JosephCarrino/mpNewton.

This formula expresses the definition of $\psi$ numerically: it is the difference (i.e., the error) between the gradient computed in exact arithmetic and the gradient computed in finite precision. However, this same formula does not apply to the case in which $\pi_g = $ np.float128, as it would result in $\psi = 0$. In this case, a slight modification is added, as follows,

$$\psi = \frac{\|\text{np.float128}(g_X(\theta^*)) - \text{np.float64}(g_X(\theta^*))\|}{u_{64}} u_{128}, \quad (3.29)$$

where $u_{128}$ and $u_{64}$ are respectively the unit roundoffs of np.float128 and np.float64.

### 3.2.1 Experiments

First, we want to prove the soundness of the results on *vanilla* Mixed-Precision Newton; thus, the step computation linear system will be solved directly using *LU* factorization, and the second-order derivatives matrix $H$ will be computed exactly. For the first experiment, the function has been chosen not to be too complicated, to take care of respecting problem bounds,

$$F_x(\theta) = \theta_1 e^{\theta_{(0)}} x + \theta_1 e^{\theta_{(1)}} x^2, \quad (3.30)$$

denoting $\nabla f$ as $g$ and $\nabla^2 f$ as $H$, following notation of previous sections. With $\theta_{(i)}$ we refer to the $i$-th dimension of $\theta$. We will perform experiments using different combinations of precisions on the same problem, starting from different points. We set $\theta^* = (6 \times 10^{-7}, 2.012 \times 10^{-2})$ and number of regression datapoints $N = 50$. For this first experiment, we ensure that $u_l \kappa(H^*) \leq 1/8$ (see *Assumptions* 3.5 and 3.8). In fact, we have $\kappa(H^*) \approx 1.8 \times 10^4$. Additionally, we show convergence curves for different starting points $\theta_0$

Figure 3.1: $\pi_g = 128, \pi_w = 64, \pi_l = 32$

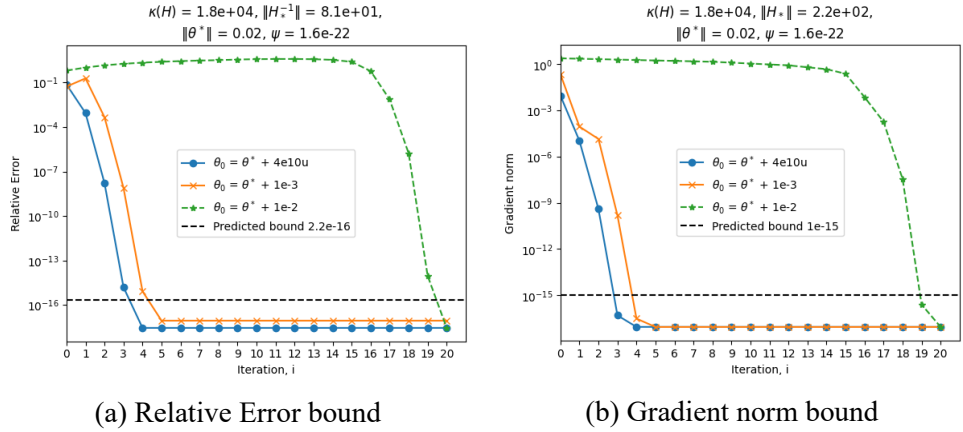

(a) Relative Error bound (b) Gradient norm bound

Figure 3.2: $\pi_g = 128, \pi_w = 64, \pi_l = 64$

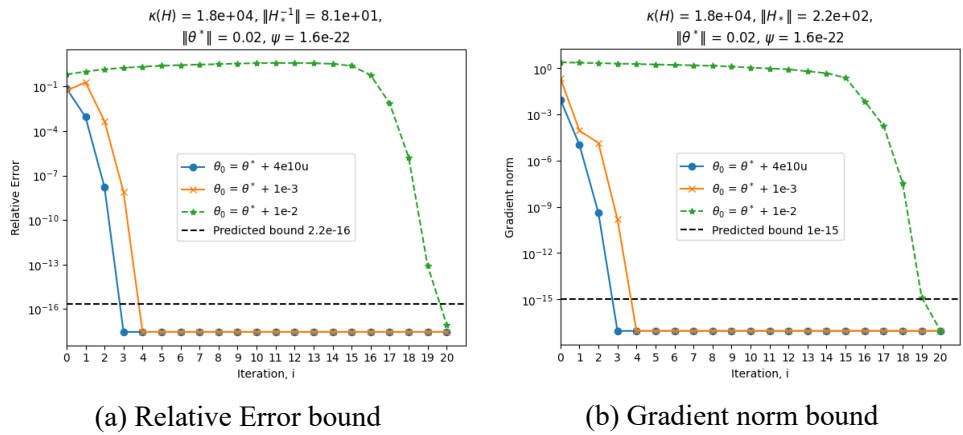

(a) Relative Error bound (b) Gradient norm bound

Figure 3.3: $\pi_g = 64, \pi_w = 64, \pi_l = 32$

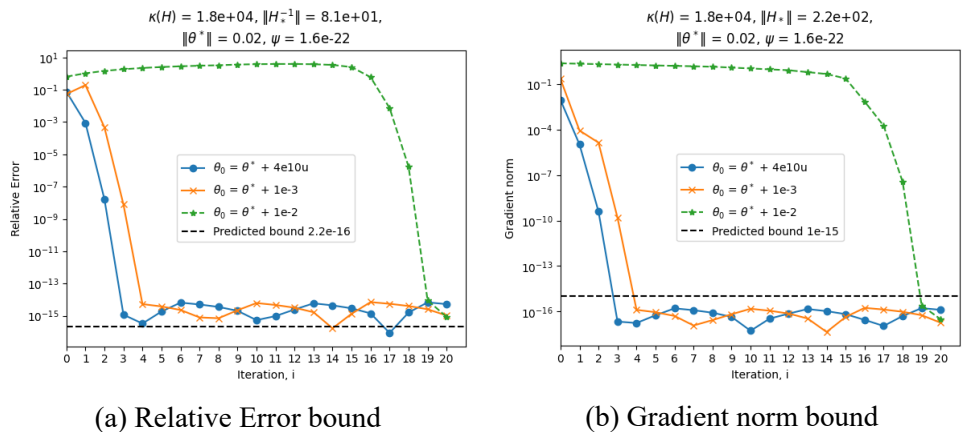

(a) Relative Error bound (b) Gradient norm bound

Figure 3.4: $\pi_g = 64, \pi_w = 32, \pi_l = 32$

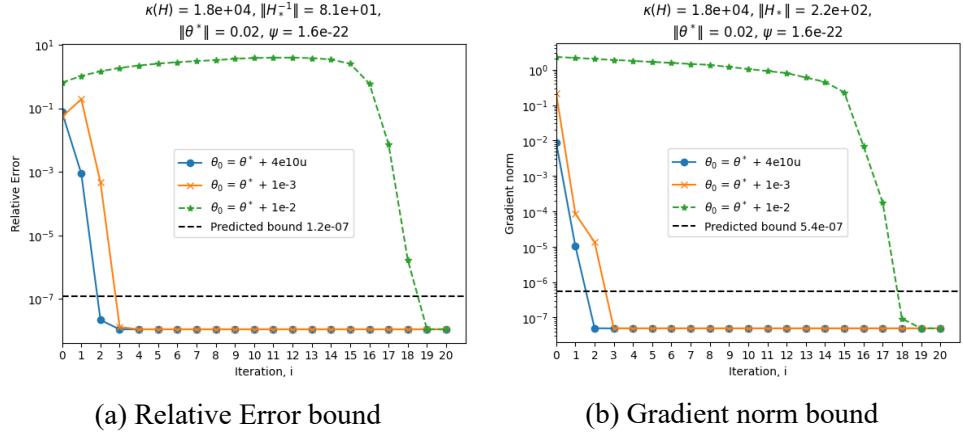

(a) Relative Error bound

(b) Gradient norm bound

The main conclusions that can be derived from this first experiment are that:

- practice respects theoretically derived bounds, confirming the soundness of our implementation;

- the bounds on both gradient norm and relative error change significantly according to working precision $\pi_w$ (see *Figure* 3.3, 3.4);

- having a very low $\psi$ value (i.e., a problem not very error-prone), the bounds are mainly influenced by $u_w$, and by $\|H\|$;

  - for relative error, since also $\|H_*^{-1}\|$ is quite low, the limiting accuracy is approximately equal to the rounding error of finite precision,

  - for gradient norm, the bound is approximately equal to $u_w$ multiplied by $\|H_*\|$.

- Since *Assumption* 3.6 is not computable due to the $L_H$ term, it is hard to define a convergent initial $\theta_0$: in the experiments, we see how all the tried starting points converge, but convergence speed is highly impacted by starting point distance, as predicted by theory.

A second experiment is conducted using a $D$-dimensional function $F$, with a diagonal Hessian that allows us to control its conditioning number easily, and hence, the compliance with the $u_l \kappa(H)$ upper bound.

$$F_x(\theta) = \theta_{(0)} + \theta_{(1)}x^2 + \sin(\theta_2 x) + ... + \sin(\theta_i x) + \theta_{(i+1)} x^{k_i}, \quad (3.31)$$

where $k_i = (i+1)//2 + 1$ and $i$ is *even*. As before, $\theta_{(i)}$ refers to the $i$-th dimension of $\theta$. Here, by using $\theta^* \approx (1, 2, 1, 2)$, we have $\kappa(H^*) = 1.4 \times 10^1$, being well below the bound on $u_l \kappa(H)$. To the solution $\theta^*$, random decimal numbers are added, to better resemble a real scenario: this is done because, by simply using integer values, the representation error coming from finite precision is equal to zero.

Figure 3.5: $\pi_g = 64, \pi_w = 32, \pi_l = 32$

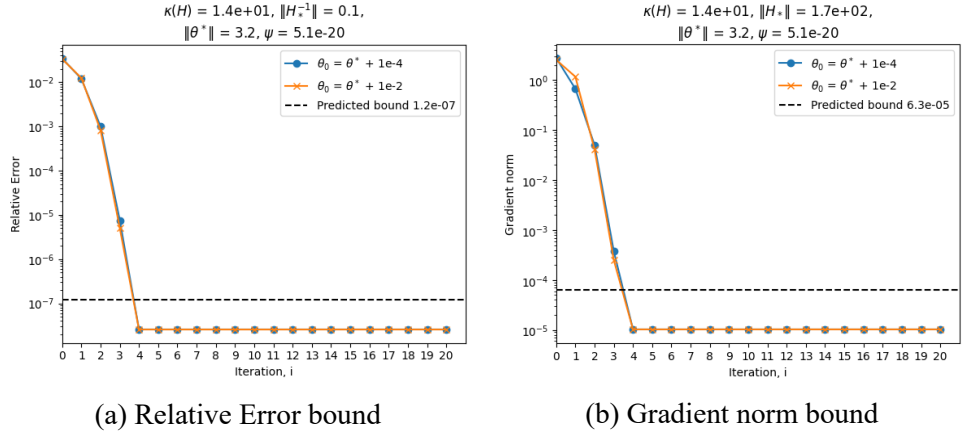

(a) Relative Error bound  (b) Gradient norm bound

We can see how, by respecting the theorem assumptions and using a more nonlinear problem, both bounds are respected *and* sharp. Thanks to the structure of the function, we can change the solution to be $\theta^* \approx (1, 2, 1e2, 2)$, to obtain a significantly higher conditioning number on $H$, but still respect the bound.

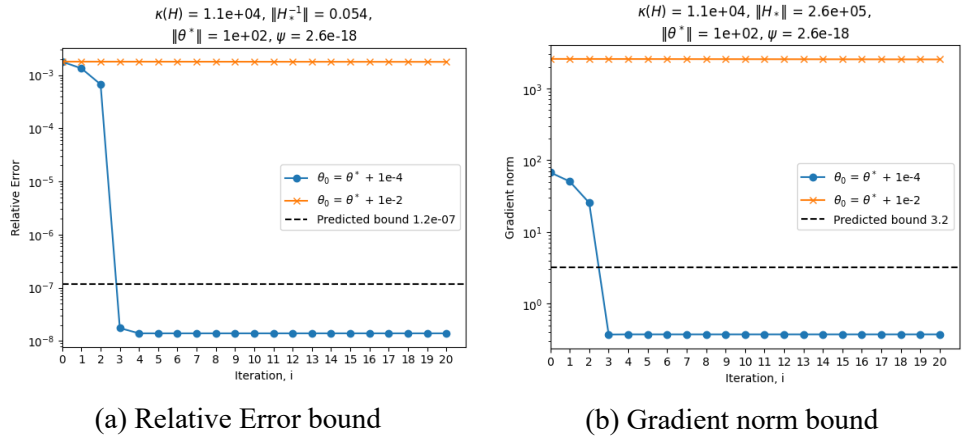

Figure 3.6: $\pi_g = 64, \pi_w = 32, \pi_l = 32$

(a) Relative Error bound

(b) Gradient norm bound

We can see how the very high conditioning number, along with a highly non-linear function, significantly impacts the gradient norm bound; we are practically not even converging to a stationary point, even though we are incredibly close to $\theta^*$. However, since our bounds on the conditioning number are still respected, we observe that the minimization curves fall below the limiting accuracy before stopping if we start close enough to our solution. This does not happen for the second starting point, due to a probably higher $L_H$ in the minimum region.

**Inexact Newton**

Regarding the analyses on the Inexact Newton method, seen in *Section* 3.1.1, some tests have also been performed. In particular, different stopping conditions $\eta$ have been tested for the iterative solver, compared with the derived bound of $1/7$. To do so, the step computation linear system is solved using the **Conjugate Gradient** method [19] - from now on referred to as **CG**. The chosen model function is the same as the last section (see *Eq.* 3.31). First, we use a high maximum number of iterations ($\sim 50$), and we set $\eta = u_l$. Comparing this setting with a Newton method that uses a direct method as the solver, we should see no difference at all. In fact, with iterative solving, we are reaching an error that is the minimum possible, i.e., the one inherent to finite representation, which is unavoidably present also when using direct methods for solving the linear system.

Figure 3.7: $\eta = u_l, \pi_g = 64, \pi_w = 64, \pi_l = 32$

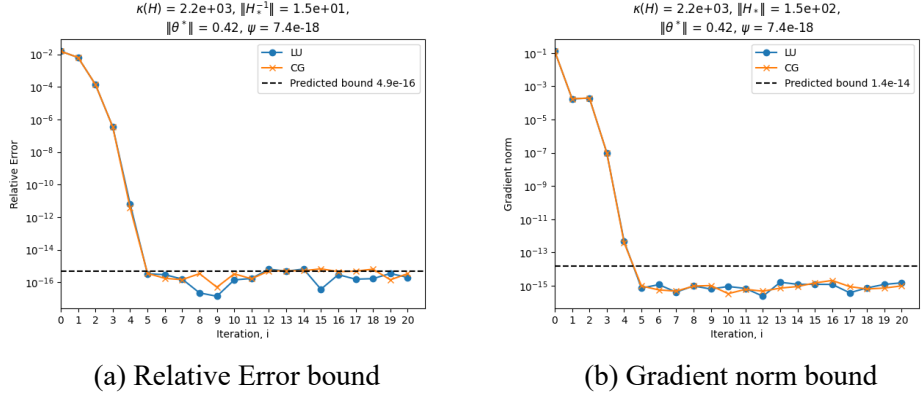

(a) Relative Error bound  (b) Gradient norm bound

As expected, no difference is visible between Newton and Inexact Newton if a very low $\eta$ is chosen. Now we want to focus on the results derived in *Section* 3.1.1: the fact that choosing a $\eta \leq 1/7$, attainable relative error and gradient norm of Mixed-Precision Newton should not differ from using a direct method.

Figure 3.8: $\eta = 1/7, \pi_g = 64, \pi_w = 64, \pi_l = 32$

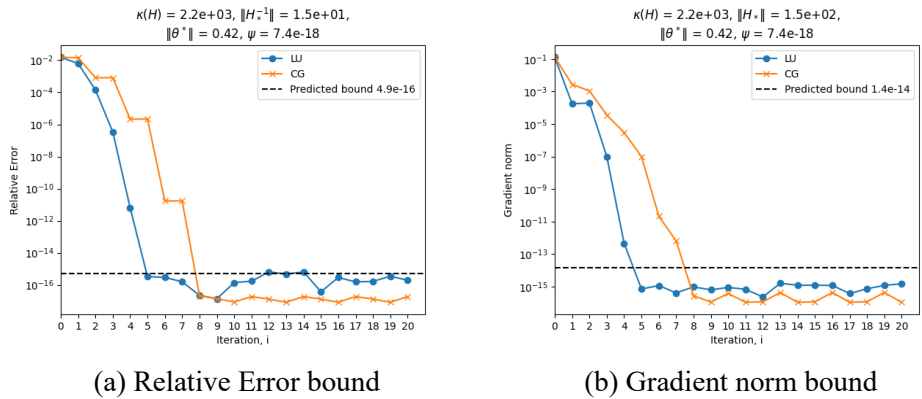

(a) Relative Error bound  (b) Gradient norm bound

As our theory predicts, the method converges until $lim_{acc}$ and $lim_g$ are reached. What differs from the usage of a direct method is solely the convergence rate, which is negatively impacted by the linear system solving error.

**Quasi-Newton**

We also tested the soundness of our Mixed-Precision Gauss-Newton algorithm (see 3). This has been done on the same D-dimensional function described in *Eq.* 3.31, with the same setting as our Newton's experiment. In this case, it must be checked that the discarded value $S(\theta)$ is lower than a certain threshold: for testing purposes, we have checked it to be lower than the smallest eigenvalue $\lambda$ of $J(\theta^*)^T J(\theta^*)$. Note that this is needed to respect the assumptions on Gauss-Newton convergence, and it does *not* impact the attainable accuracy; however, a significant value $S(\theta)$ can slow the convergence significantly.

Figure 3.9: $\pi_g = 128, \pi_w = 64, \pi_l = 32$

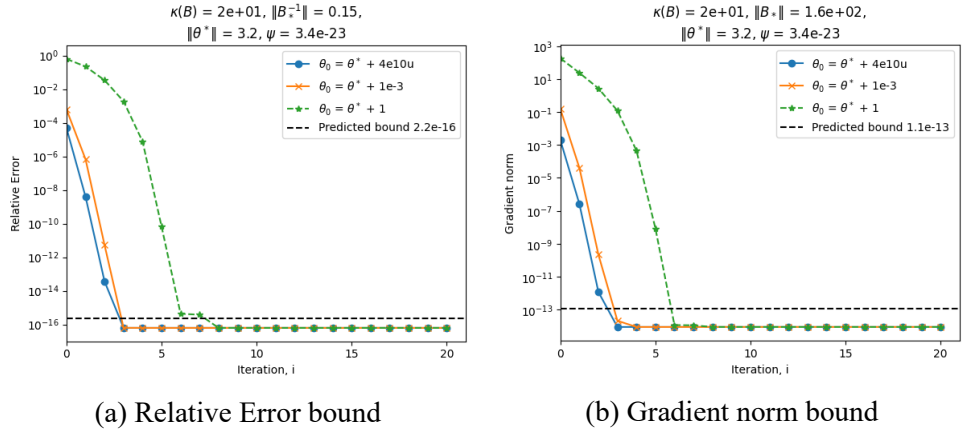

(a) Relative Error bound        (b) Gradient norm bound

Figure 3.10: $\pi_g = 128, \pi_w = 64, \pi_l = 64$

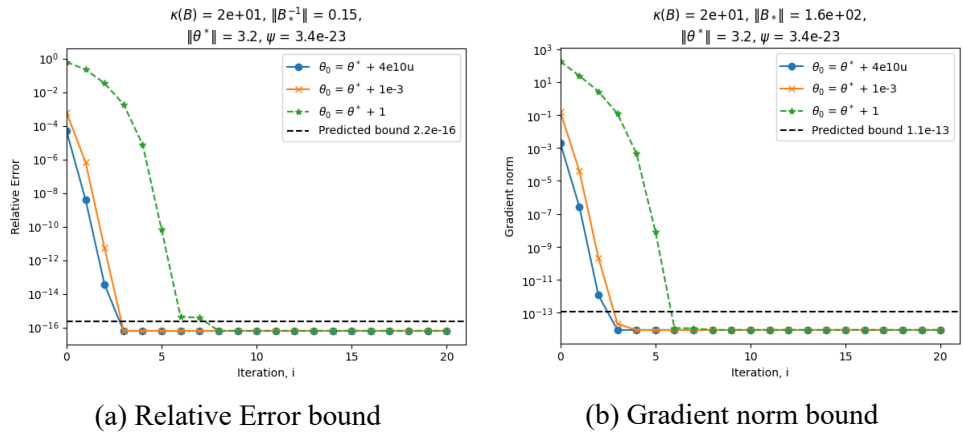

(a) Relative Error bound        (b) Gradient norm bound

Figure 3.11: $\pi_g = 64, \pi_w = 64, \pi_l = 32$

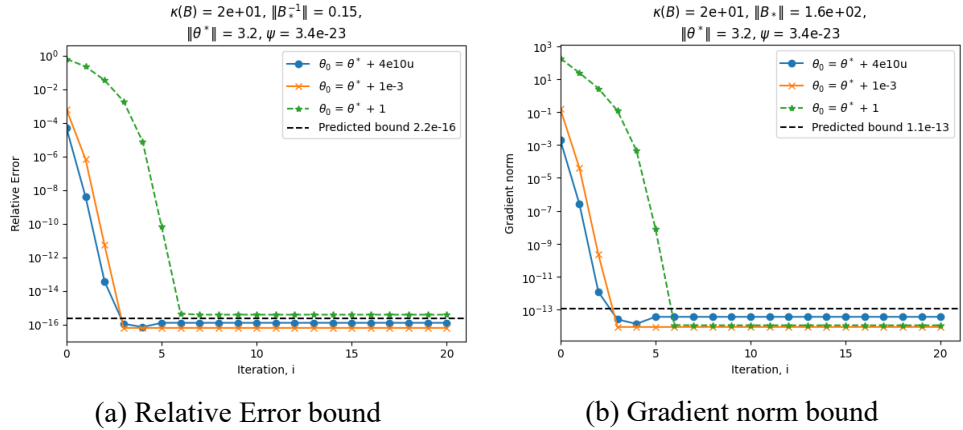

(a) Relative Error bound

(b) Gradient norm bound

Figure 3.12: $\pi_g = 64, \pi_w = 32, \pi_l = 32$

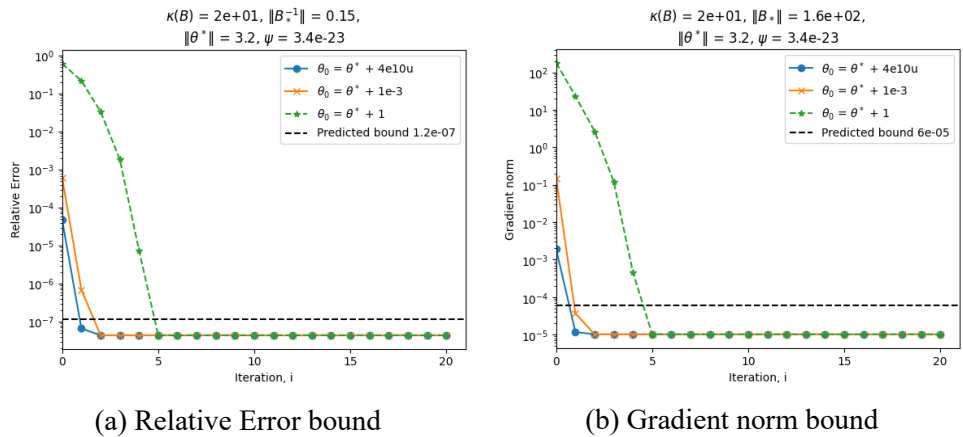

(a) Relative Error bound

(b) Gradient norm bound

It is notable from the charts that the $lim_{acc}$ and $lim_g$ bounds are respected from different starting points, also using Mixed-Precision Gauss-Newton approximation. Checking the discarded $S(\theta)$ term, we have assessed that the upper-bound on it is respected for all starting points, except $\theta_0 = \theta^* + 1$: in this case, as theory predicts, the attained accuracy is still the same, but the convergence rate is significantly impacted.

## 3.3 Coupling with gradient conditioning

The purpose of this section is to focus on the term $\psi$ of our convergence theorems 3.1.1 and 3.1.2. As we saw both analytically and practically in the last section, it is the term that primarily impacts the attainable accuracy of our minimization method. For the previous experiments, this is a consequence of finite-precision computation of the gradient; however, we expect our model to be general enough to include other kinds of approximations in $\psi$ as well.

### 3.3.1 Finite-Differences

An example of gradient approximation for optimization is finite differences, which is a widely used technique for zeroth-order optimization [39]. The most stable finite-difference computation is central differences, which are used to compute the gradient as follows

$$\hat{g} = \frac{f(\theta + \epsilon) - f(\theta - \epsilon)}{2\epsilon}, \tag{3.32}$$

where $\mathbb{R} \ni \epsilon \ll 1$. The choice of $\epsilon$ is of fundamental importance. A simple intuition is that it cannot be $\epsilon < u_g$, because it would be represented as 0. However, the choice of $\epsilon = u_g$ is not the most suitable: this is well explained in Nick Higham's blogpost on the topic[4]. The article points out that:

- a too high $\epsilon$ leads to bad approximation of the gradient, for the definition of the gradient itself, which implies $\epsilon \to 0$;

- since we are in a finite-precision context, a too small $\epsilon$ can lead to *cancellation errors*.

The suggested approach is to choose $\epsilon \approx u_g^{1/2}$, to minimize both approximation and finite-precision errors. Given the definition of our approximated $\hat{g}$, we now want to assess that the $lim_{acc}$ and $lim_g$, which depend on $\psi$, are respected

---

[4]https://nhigham.com/2020/10/06/what-is-the-complex-step-approximation/

when using a finite-difference gradient. Given this setting, $\psi$ is computed as

$$\psi = \|\text{np.float128}(g_X(\theta^*)) - \pi_g(\hat{g}_X(\theta^*))\|, \quad (3.33)$$

i.e., including the finite-difference error in the term. For our experiment, we will use the same model function as defined in *Eq.* 3.31, and the same setting used for the previous section's experiments.

Figure 3.13: Finite-Difference $\hat{g}$, $\pi_g = 128, \pi_w = 64, \pi_l = 32$

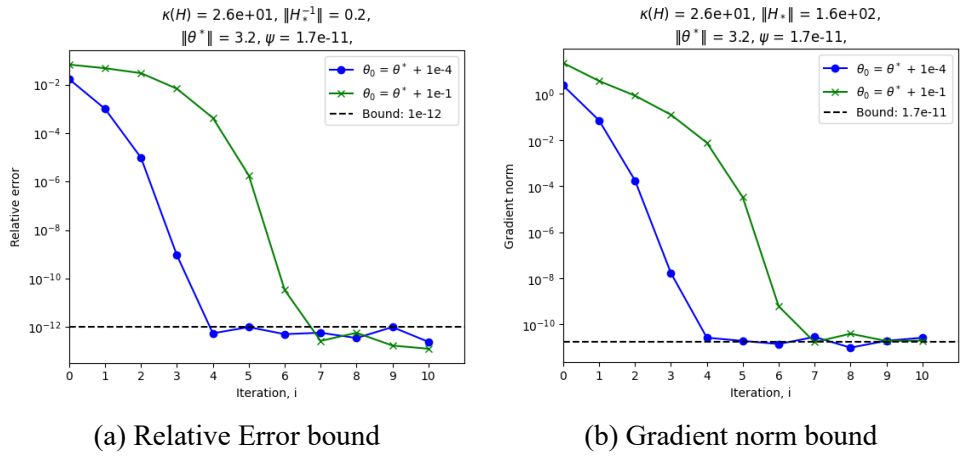

(a) Relative Error bound    (b) Gradient norm bound

Figure 3.14: Analytical $g$, $\pi_g = 128, \pi_w = 64, \pi_l = 32$

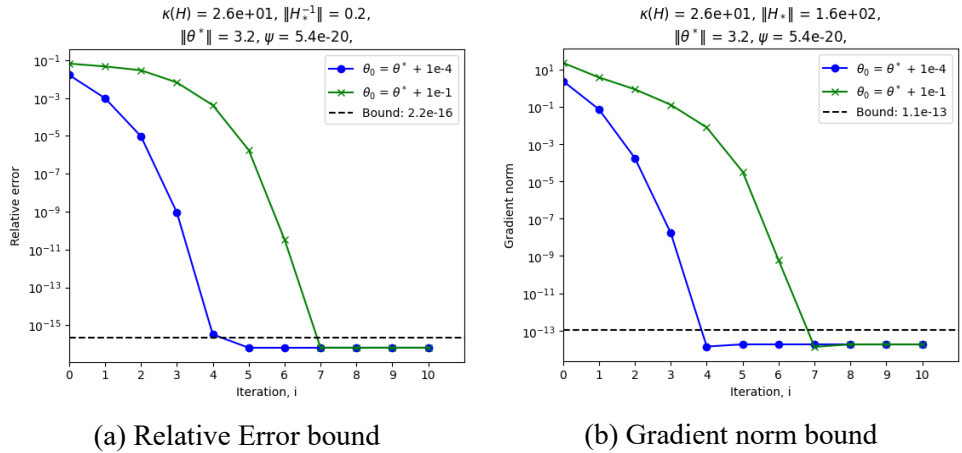

(a) Relative Error bound    (b) Gradient norm bound

For *Figure* 3.13, we have a $\psi \approx 1.5 \times 10^{-11}$, significantly higher than the one computed with exact $g$, and showing a very sharp bound. Also for analytical $g$, as in previous experiments, the bounds are respected, both being sharp. We observe how the usage of a finite-difference $\hat{g}$ has been taken into account in the bounds and reflected in the numerical results, thereby proving the soundness of our framework also for cases in which gradient approximations are not solely due to finite precision. With this same experimental setting, we can also support experimentally Nick Higham's blog post intuitions regarding the choice of the best $\epsilon$. In the following charts, Mixed-Precision Newton's convergences are compared using different $\epsilon$, and, for each choice of it, different $\psi$ are computed. For each chart, a different $\pi_g$ is chosen to show that the selection of $\epsilon$ is highly dependent on the chosen precision.

Figure 3.15: $\pi_g = 128, \pi_w = 64, \pi_l = 32$

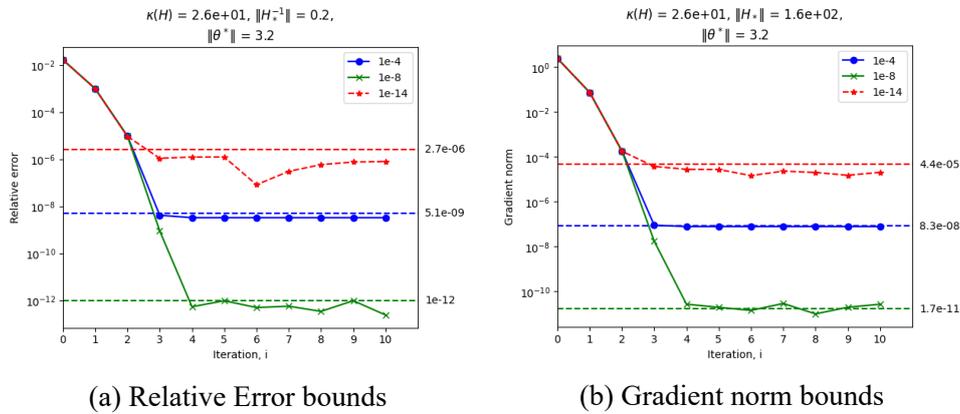

(a) Relative Error bounds  (b) Gradient norm bounds

Figure 3.16: $\pi_g = 64, \pi_w = 64, \pi_l = 32$

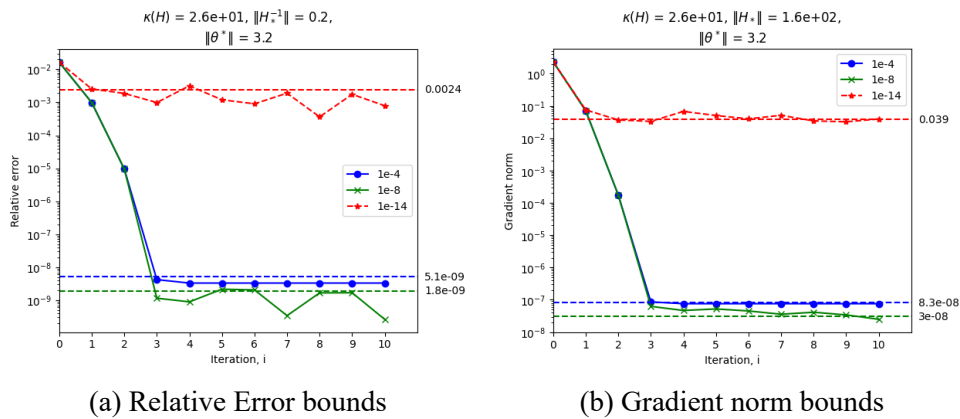

(a) Relative Error bounds  (b) Gradient norm bounds

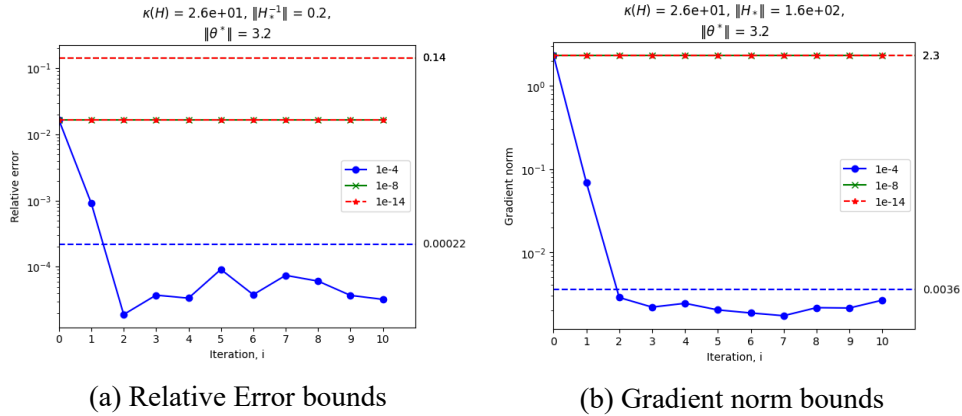

Figure 3.17: $\pi_g = 32, \pi_w = 64, \pi_l = 32$

(a) Relative Error bounds

(b) Gradient norm bounds

As expected, we see how for all the cases in which $\epsilon < u_g$ there is no convergence of Newton's method: see, for example, $\epsilon = 1 \times 10^{-14}$ in *Fig.* 3.17, where $\pi_g = 32$ and $u_g \approx 1 \times 10^{-6}$. We also see how the choice of $\epsilon \approx u_g^{1/2}$ is always the fastest and most reliable one, even though, in some cases, it shows few differences from other options (see *Fig.* 3.16).

### 3.3.2 Ill-conditioned gradient

In the previous section, we focused on a specific approximated gradient and showed how to deal with finite differences in a finite-precision context. However, we would like to demonstrate how, thanks to our framework, it is possible to tune the precision $\pi_g$ to address gradients that are inherently error-prone, thereby proving that using a higher precision for the gradient can improve our convergence without the need to use high precision for the whole minimization algorithm. To do so, we first need to understand what an error-prone gradient computation is and what it depends on. We can see $\psi$ value as a backward-error on $g$, and it is formalized as the *conditioning* of the gradient, from now on denoted as $\kappa(g)$. In relative terms,

$$\kappa(g(x)) = \frac{\|g(x + \Delta x) - g(x)\|/\|g(x)\|}{\|\Delta x\|/\|x\|} \quad \text{where } \|\Delta x\| \leq \|x\|, \quad (3.34)$$

i.e., $k(g(x))$ indicates how much does $g$ changes with a small change on the input $x$. Doing some derivations, we can see the dependence of this value on the second-order derivative of $f$, i.e., the Hessian $H$, as follows,

$$\begin{aligned}
\kappa(g(x)) = &= \frac{\|g(x+\Delta x) - g(x)\|}{\|\Delta x\|} \cdot \frac{\|x\|}{\|g(x)\|} \\
&= \frac{\|g(x+\Delta x) - g(x) - H(x)\Delta x + H(x)\Delta x\|}{\|\Delta x\|} \cdot \frac{\|x\|}{\|g(x)\|} \\
&\leq \left( \frac{\|g(x+\Delta x) - g(x) - H(x)\Delta x\|}{\|\Delta x\|} + \frac{\|H(x)\Delta x\|}{\|\Delta x\|} \right) \cdot \frac{\|x\|}{\|g(x)\|} \\
&\underbrace{\leq}_{A.0.1} \left( \frac{L_H}{2} \|\Delta x\| + \|H(x)\| \right) \cdot \frac{\|x\|}{\|g(x)\|}.
\end{aligned} \quad (3.35)$$

This means that:

- the gradient conditioning tends to infinity when close to the solution ($\|g(x)\| \to 0$);

- a badly conditioned gradient can mean a Lipschitz-discontinuous Hessian.

This tells us that, if we have a function whose gradient is ill-conditioned, it is possible that a sequence generated from a starting point $x_0$ far from the solution $x^*$ could not converge as expected, in a finite precision context (see *Assumption* 3.6 of our theorems). We now want to test our Mixed-Precision Newton method, which minimizes a function whose gradient is ill-conditioned. This is not trivial, as we still want to keep the Hessian conditioning low enough. To experiment with this, we defined a 2-dimensional function $f$ to be minimized, with particular parameters that allow us to control both $\kappa(H)$ and $\kappa(g)$ - almost - distinctively:

$$f(\theta) = \sqrt{1 + \alpha_x(\theta_0 - a)^2} + \sqrt{1 + \alpha_y(\theta_1 - b)^2}, \quad (3.36)$$

where $\alpha_x, \alpha_y \in \mathbb{R}$ are hyperparameters that allow us to control conditioning numbers, and $a, b \in \mathbb{R}$ are the solution (i.e. $\mathbb{R}^2 \ni \theta^* = (a, b)$). We can easily see how the gradient gets more ill-conditioned by using $\alpha_x, \alpha_u \gg 1$, and that $k(H^*) = \alpha_x/\alpha_y$. Hence, to have a high gradient conditioning keeping $k(H^*)$ under control, we must choose $\alpha_x \approx \alpha_y \gg 1$.

We choose $\theta^* = (a, b) \approx (1 \times 10^{-7}, 2 \times 10^{-8})$. For the first experiment, we choose $\alpha_x \approx 5 \times 10^2$ and $\alpha_y \approx 5 \times 10^{-4}$, so that we have a high $\kappa(H^*)$ - in fact we have $u_l \kappa(H^*) \approx 1/8$ which is our theorems bound. With this choice of $alpha_x$ and $\alpha_y$, we also still have good gradient conditioning, as seen in the next charts, which look at the low values of $\psi$.

Figure 3.18: low $\psi, \pi_g = 64, \pi_w = 64, \pi_l = 32$

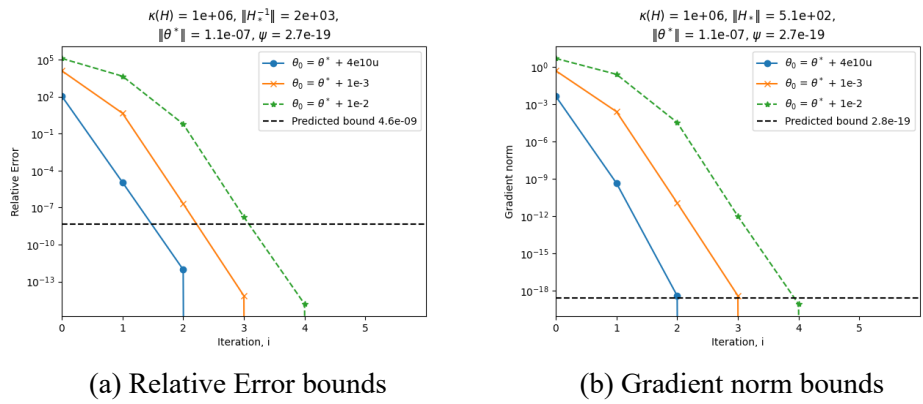

(a) Relative Error bounds

(b) Gradient norm bounds

Figure 3.19: low $\psi, \pi_g = 32, \pi_w = 64, \pi_l = 32$

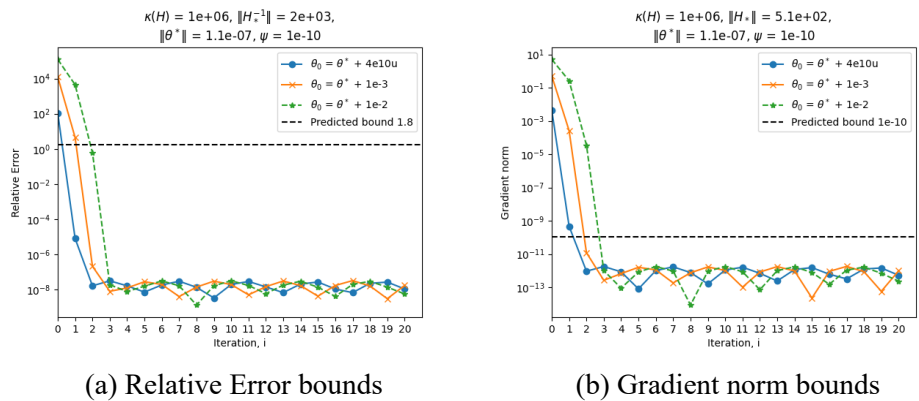

(a) Relative Error bounds

(b) Gradient norm bounds

We now significantly decrease the value of $\alpha_x$ and $\alpha_y$ to achieve a much higher $\kappa(g)$ and examine how this relates to different choices of $\pi_g$. To do so, we choose $\alpha_x \approx 5 \times 10^1$ and $\alpha_y \approx 5 \times 10^7$, achieving the same $\kappa(H^*)$ as before, but a higher error on the gradient computation.

Figure 3.20: high $\psi$, $\pi_g = 64$, $\pi_w = 64$, $\pi_l = 32$

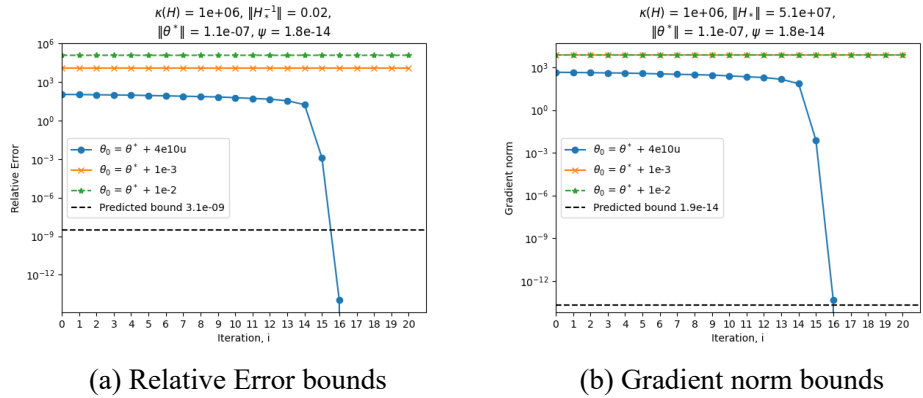

(a) Relative Error bounds    (b) Gradient norm bounds

Figure 3.21: high $\psi$, $\pi_g = 32$, $\pi_w = 64$, $\pi_l = 32$

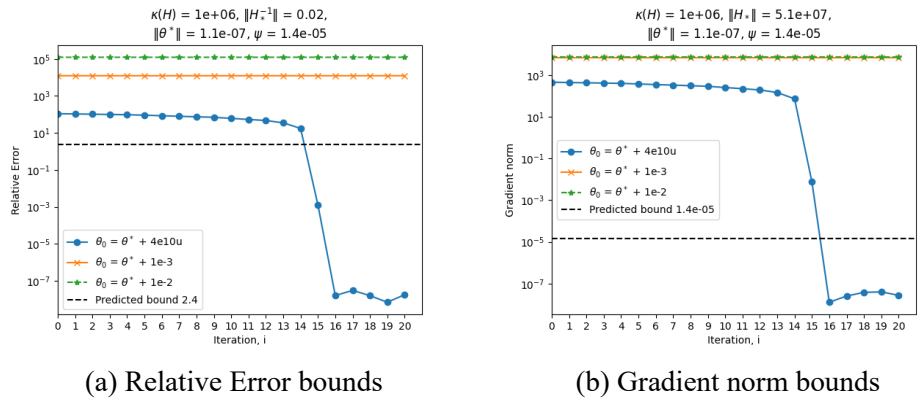

(a) Relative Error bounds    (b) Gradient norm bounds

We observe that the $\psi$ of our function exhibits a significantly higher value, which is reflected in the higher limiting accuracy values. Also, we see how, by increasing $\psi$, the $\theta_0$ starting further from the solution are unable to converge, due to a probably higher $L_H$ (see (3.35) and *Assumption* 3.6). In conclusion, it appears that, at least from this experiment, tuning $\pi_g$ does not enable convergence with starting points that would not converge in lower precision; however, it significantly improves the quality of the achieved solutions.

## 3.4 Classification experiments

As extensively shown in the Background section (see 2.3), it is possible to use Newton's method to minimize a loss function and train Machine Learning models, as is usually done with first-order optimizers, such as Adam. In this section, we will focus on the training of a logistic regressor, minimizing the Binary Cross-Entropy loss defined in (2.9), on different datasets from the UCI repository[5]. However, Newton's method requires a positive definite Hessian to converge, which is not guaranteed by the classic BCE loss. To overcome this, we have used a **L2**-regularization, as follows

$$L(y, \hat{y}, \theta) = BCE(y, \hat{y}) + \frac{\lambda}{2}\|\theta\|^2, \qquad (3.37)$$

where $\mathbb{R} \ni \lambda \ll 1$ is an hyperparameter (chosen to be $\approx 1 \times 10^{-4}$) and $\theta$ are the parameters to estimate. Given this loss function, we have chosen two datasets, Australian [38] and MUSH [46], to train a logistic regressor on, using our Mixed-Precision Newton. For these experiments, *LU* factorization is used for linear system solving, and the true Hessian is computed, thereby reducing approximation errors to those resulting from finite precision only. To compute our solution $\theta^*$, a run of 50 iterations using Newton's method in maximum precision (np.float128) is performed, and the estimated parameters are used as $\theta^*$. Firstly, we will show the results obtained on the Australian dataset, comparing Newton's method and an implementation of AdamW [33]. This latter optimizer has been tested using different precisions for all operations, with no significant differences observed. Hence, in the tests, AdamW works in working precision $\pi_w$.

We see by the charts in *Figure* 3.22 how Mixed-Precision Newton not only outperforms Adam, but also respects the boundaries given in *Theorems* 3.1.1 and 3.1.2. We also want to assess whether this leads to an overfitted model; we do so by computing confusion matrices on the test set for the tested models.

---

[5]https://archive.ics.uci.edu/datasets

Figure 3.22: Australian dataset, $\pi_g = 128, \pi_w = 64, \pi_l = 32$

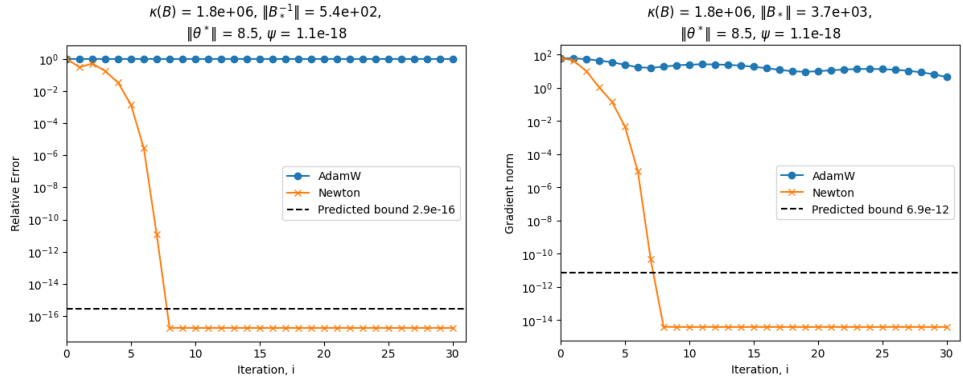

(a) Relative Error  (b) Gradient norm bound

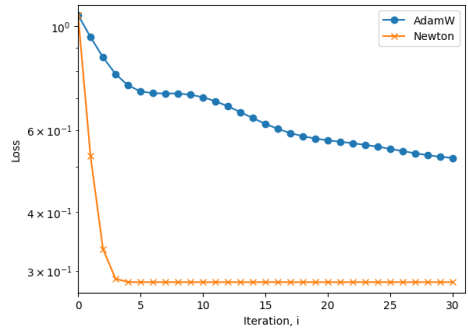

(c) Loss

To calculate the predicted classes, a threshold of 0.5 is applied to the output logits.

The results of Newton's method are significantly better than those of AdamW, as seen in the confusion matrices of *Figure* 3.23 on the test set.

Since the usage of iterations as a term of comparison between the two methods can be unfair to Adam, because each of its iterations requires significantly fewer computations than Newton's method, we also plotted the same charts in a time-wise manner, in *Figure* 3.24. It is worth noting that the tests were conducted on a personal computer, which is not capable of exploiting low precision to speed up computations. We can see how, at the same instant, Newton's shows better convergence than Adam on the Australian dataset. We now replicate the same experiment on the MUSH dataset, which is also known to be

Figure 3.23: Australian dataset, $\pi_g = 128, \pi_w = 64, \pi_l = 32$

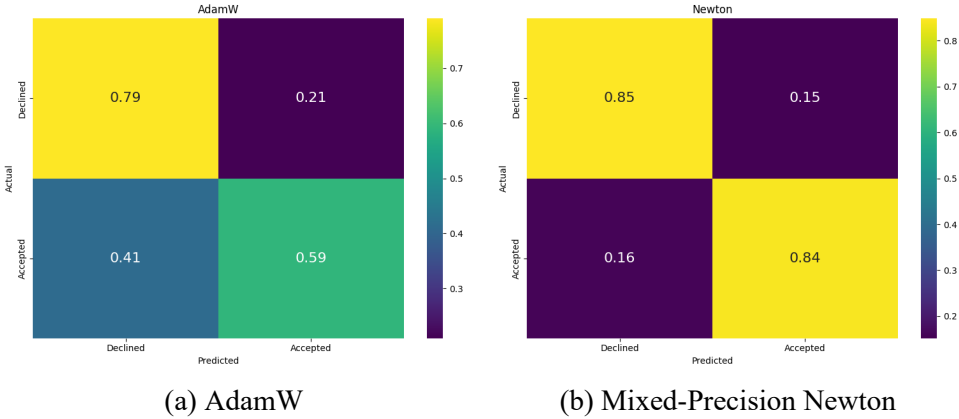

(a) AdamW  (b) Mixed-Precision Newton

hardly learnable with small machine learning models, such as a Logistic Regressor, if using first-order optimizers. *Table* 3.1 shows the results on MUSH dataset of both AdamW and Mixed-Precision Newton's optimizers. The gradient norm and loss function values reported are the ones achieved after one second of running, whereas the "True-Positives" (TP) and "True-Negatives" (TN) values refer to the results achieved on the test set. Even though showing good results when using Mixed-Precision Newton's method, the loss function defined in 3.37 is not in a Least-Squares form, making it impossible to use Gauss-Newton optimizer to minimize it. To test also this method on a regression task, we used a loss function called *Square-Error* loss, defined as follows:

$$L_{LS}(\theta) = \frac{1}{2N} \sum_{i=1}^{N} (y_i - \sigma(x_i \cdot \theta))^2 + \frac{\lambda}{2} \|\theta\|^2. \quad (3.38)$$

| Optimizer | Gradient Norm | Loss value | TN | TP |
|---|---|---|---|---|
| **AdamW** | $10^0$ | $6 \times 10^{-1}$ | 0.94 | 0.60 |
| **Newton** | $10^{-16}$ | $2 \times 10^{-1}$ | 0.97 | 0.94 |

Table 3.1: Logistic Regressor - MUSH

This function has been minimized with respect to both tested datasets, using AdamW, Mixed-Precision Newton, and Mixed-Precision Gauss-Newton. The results are summarized in the following *Tables* 3.2 and 3.3. As in the previous table, the reported relative error and gradient norm values are those achieved

Figure 3.24: Australian dataset - timewise, $\pi_g = 128, \pi_w = 64, \pi_l = 32$

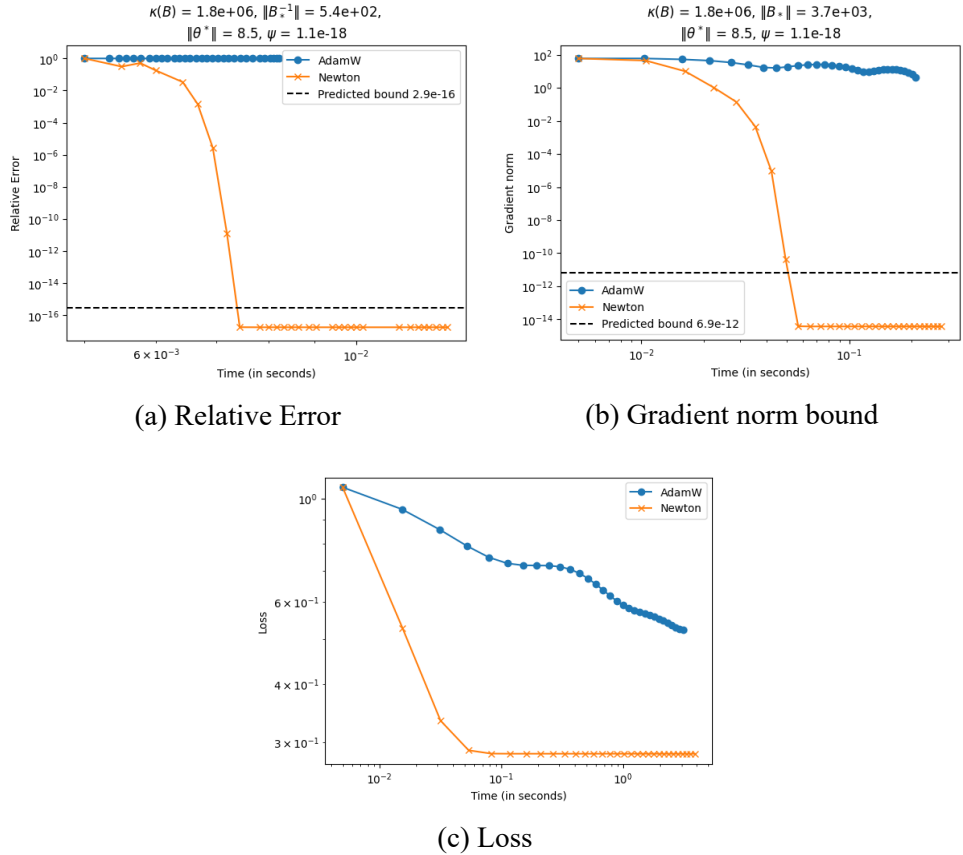

(a) Relative Error

(b) Gradient norm bound

(c) Loss

by the optimizers after one second. The "True-Positives" (TP) and "True-Negatives" (TN) values refer to the predictions of the trained regressors on the test set. The used precisions are the same as the previous experiment, being $\pi_g = 128, \pi_w = 64$, and $\pi_l = 32$.

| Optimizer | Gradient Norm | Loss value | TN | TP |
|---|---|---|---|---|
| **AdamW** | $10^2$ | $1 \times 10^{-1}$ | 0.85 | 0.50 |
| **Newton** | $10^1$ | $3 \times 10^{-2}$ | 0.65 | 0.75 |
| **Gauss-Newton** | $10^{-1}$ | $2 \times 10^{-2}$ | 0.82 | 0.85 |

Table 3.2: LeastSquares Regressor - Australian

We can see that Gauss-Newton performs the same as Newton, or better. We also report the loss function chart over time in *Figure* 3.25: as expected, Gauss-Newton is significantly faster than Newton. On the other hand, we see how Newton's method performs worse using *Square-Error* loss than with the classical Binary-Cross Entropy, for reasons to be investigated.

| Optimizer | Gradient Norm | Loss value | TN | TP |
|---|---|---|---|---|
| **AdamW** | $10^2$ | $1 \times 10^{-1}$ | 0.94 | 0.68 |
| **Newton** | $10^1$ | $6 \times 10^{-2}$ | 0.97 | 0.99 |
| **Gauss-Newton** | $10^{-1}$ | $4 \times 10^{-2}$ | 0.97 | 0.99 |

Table 3.3: LeastSquares Regressor - MUSH

Figure 3.25: Least-Squares regressor loss over time - MUSH

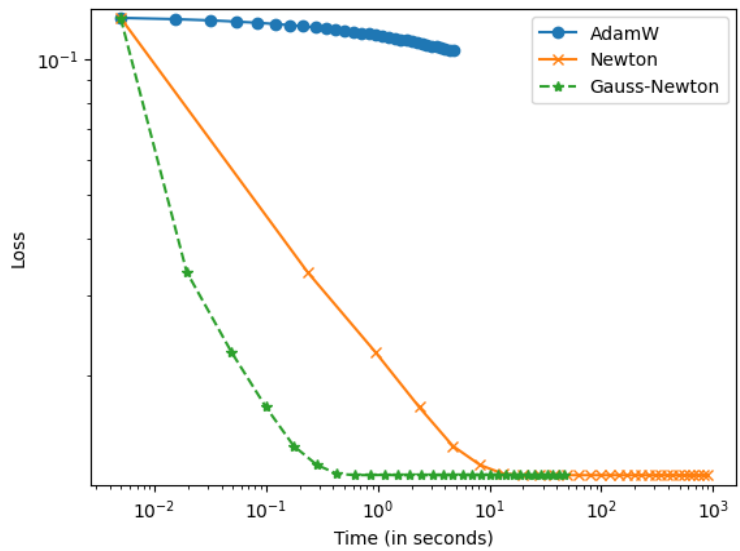

# Chapter 4

# Gauss-Newton_k

In the previous chapter, we have seen how Mixed-Precision Gauss-Newton has shown remarkable performance, with a lower computational cost, and keeping the convergence properties of Theorems 3.1.1 and 3.1.2. In some cases, however, the residuals we would like to minimize can be very nonlinear; with such functions, it is not possible to discard second-order information and converge as rapidly as Newton's method. Keeping the regression setting of *Eq.* (2.7), one example of a highly non-linear model function is the following

$$\min_\theta f_X(\theta) = \frac{1}{2}\|F_X^{lin}(\theta) - y\|^2 + \frac{1}{2}\|F_X^{nonlin}(\theta) - y\|^2$$
$$F_X^{lin}(\theta) = (\sum_{i=1}^m \theta_i^2)AX \qquad (4.1)$$
$$F_X^{nonlin}(\theta) = \sin(\prod_{i=1}^m \theta_i X),$$

where $A \in \mathbb{R}^{n \times n}$ is randomly generated. In *Fig.* 4.1, we try to regress on the model function of *Eq.*( 4.1) with both Newton's and Gauss-Newton's methods, starting from the same $\theta_0$ and using only double precision. The chart shows how Newton's method converges quadratically, whereas Gauss-Newton seems not to converge at all. Ideally, we would like to take advantage of the low computational weight of Gauss-Newton, without sacrificing convergence in highly non-linear cases, such as the reported one.

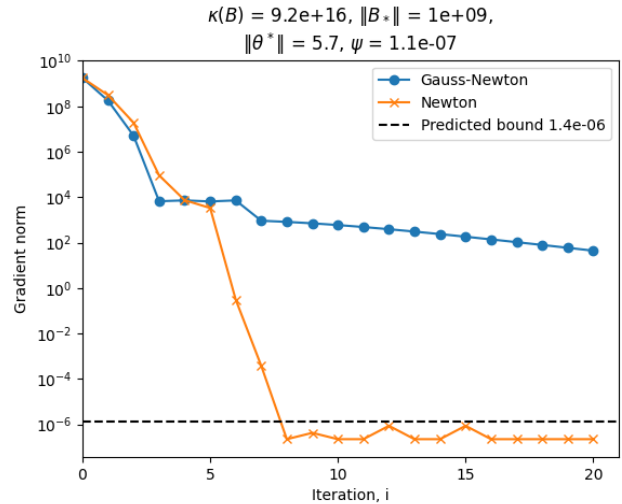

Figure 4.1: $\pi_g = 64, \pi_w = 64, \pi_l = 64$

To achieve this, we developed the idea of **Gauss-Newton_k**; it is a minimization algorithm that falls between Newton's and Gauss-Newton's methods, leveraging the structure of the term $S(x)$ defined in (3.22), by only computing the $k$ most important second-order derivatives.

## 4.1 Analytical relation with Gauss-Newton

The definition of $S(x)$ in a Least Squares problem is

$$S(x) = \sum_{i=1}^{m} r_i(x) \cdot \nabla^2 r_i(x), \qquad (4.2)$$

where $r_i$ is the $i$-th residual function of the function to minimize, whose second-order derivative is

$$\nabla^2 f(x) = J(x)^T J(x) + S(x), \qquad (4.3)$$

where $J(x)$ is the Jacobian matrix of the first order derivatives of $r_i$ for all $i \in [1, m]$. Slightly abusing notation, we can define $S_j(x)$ as the following

$$S_j(x) = r_j(x) \cdot \nabla^2 r_j(x), \qquad (4.4)$$

which means that $S(x) = \sum_j S_j(x)$. In Newton's method, we would simply use $\nabla^2 f(x)$ as in (2.3), whereas Gauss-Newton proposes to discard the term $S(x)$, neglecting the residual's second-order derivatives. The proposal of **Gauss-Newton_k** (from now on simply **GN_k**) is to compute a subset $K \subseteq [1, m]$ of the $m$ $S_j(x)$, approximating the second-order derivative of $f$ as

$$\nabla^2 f(x) \approx B(x) = J(x)^T J(x) + \sum_{j \in K} S_j(x). \tag{4.5}$$

This generic definition of the second-order derivative implies that

- if $|K| = m$, we are computing the full $S(x)$, i.e., we are using Newton's method;

- if $|K| = 0$, we are not computing $S(x)$, i.e., we are using Gauss-Newton's method;

- if $0 < |K| < m$, we are using some residuals' second-order information, discarding others.

In this setting, we can rethink the conditions of convergence of Gauss-Newton, and hence of our Mixed-Precision Newton's theorem (as explained in 3.1.2); we can define $\bar{K} \triangleq [1, m] - K$ and rewrite the condition (3.27) as the following

$$\left\| \sum_{j \in \bar{K}} S_j(x)(x - x^*) \right\| \leq \sigma \|x - x^*\|, \tag{4.6}$$

in other words, we want to upper-bound only the information we discard. This rewritten condition also works for the $|K| = 0$ and $|K| = m$ edge cases, and allows us to use *Theorems* 3.1.1 and 3.1.2 conclusions without further proofs.

What is purposely missing from this definition of GN_k is how we compute $K$, i.e., how do we select which $S_j(x)$ we should calculate. For our algorithm formulation, we propose to use the $S_j(x)$ with the bigger magnitude residuals $r_j(x)$.

Firstly, we propose the pseudo-code of GN_k in *Algorithm* 4: $k$ is a hyperparameter that defines how many $S_j(x)$ we want to compute. Then, ap-

---
**Algorithm 4** Gauss-Newton_k

    **for** $i = 1 : maxit$ or until converged **do**
        Compute $r_j(x_i)$ for all $k \in [1, m]$
        Define $K = \{j | |r_j(x_i)| \text{ is in the biggest } k\}$
        Compute $S(x_i) = \sum_{j \in K} r_j(x_i) \cdot \nabla^2 r_j(x_i)$
        Compute $g_i = J(x_i)^T R(x_i)$
        Solve $(J(x_i)^T J(x_i) + S(x_i))d_i = -g_i$
        $x_{i+1} = x_i + d_i$
    **end for**

---

plying Mixed-Precision Newton theory as in *Chapter* 3, we propose a mixed-precision version of GN_k in *Algorithm* 5 The computed $r_j(x_i)$ are then used

---
**Algorithm 5** Mixed-Precision Gauss-Newton_k

    **for** $i = 1 : maxit$ or until converged **do**
        Compute $r_j(x_i)$ for all $k \in [1, m]$ ← in high-precision $\pi_g$
        Define $K = \{j | |r_j(x_i)| \text{ is in the biggest } k\}$
        Compute $S(x_i) = \sum_{j \in K} r_j(x_i) \cdot \nabla^2 r_j(x_i)$ ← in working-precision $\pi_w$
        Compute $g_i = J(x_i)^T R(x_i)$ ← in high-precision $\pi_g$
        Solve $(J(x_i)^T J(x_i) + S(x_i))d_i = -g_i$ ← in low-precision $\pi_l$
        $x_{i+1} = x_i + d_i$ ← in working-precision $\pi_w$
    **end for**

---

to form the matrix $R(x_i)$, which is the reason why they are computed in precision $\pi_g$. The term $S(x_i)$ is instead computed in precision $\pi_w$ since it is then used to form $\nabla^2 f(x_i)$, computed in working-precision in our Mixed-Precision Newton's setting. The idea behind the algorithm is to avoid computing $\nabla^2 r_j(x_i)$ if not needed, and the usage of $r_j(x_i)$ to do so comes from two intuitions:

- **analytically**, lower $r_j(x_i)$ means lower $S_j(x_i)$, thus its impact on the second-order derivative of $f$ may be negligible;

- **geometrically**, a low $r_j(x_i)$ means that $x_i$ is close to the solution $x^*$ in the dimension $j$, thus we may avoid using second-order information to converge furtherly, and focus on other dimensions.

### 4.1.1 Experiment

The presented algorithm 5 has been tested on the same function defined in (4.1). We defined $m = d = 2$, which are the dimensions of, respectively, datapoints $X_i$ and parameters $\theta$, and $n = 1500$. The solution $\theta^*$ has been set to $\{2.012, 5.023\}$ and $\theta_0 = \theta^* + 1 \times 10^1$. From now on, the quantity $k$ will refer not to an absolute number of used $S_j(x)$, but to a *percentage*. In the following charts, we compare GN_0, i.e., Gauss-Newton, GN_100, i.e., Newton, and GN_30, i.e., GN_k where 30% of the $S_j(x)$ are chosen as the largest ones and computed, using *Algorithm* 5.

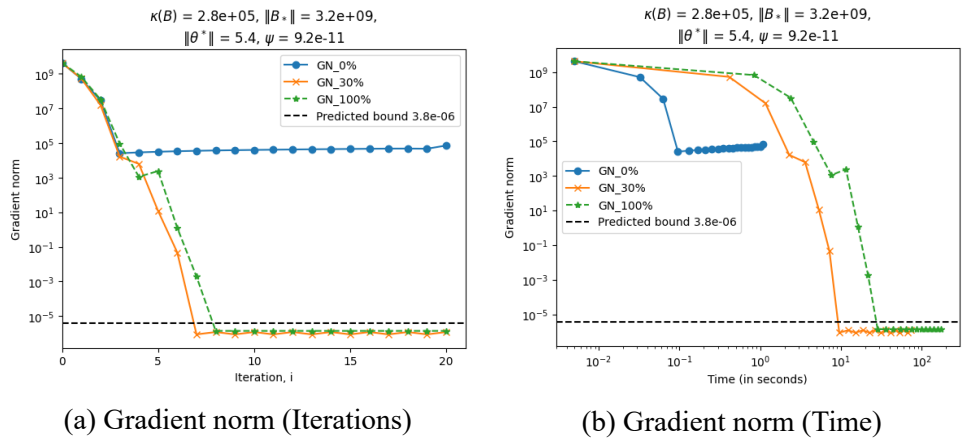

Figure 4.2: high $\psi, \pi_g = 64, \pi_w = 64, \pi_l = 64$

(a) Gradient norm (Iterations)  (b) Gradient norm (Time)

*Figure* 4.2 shows us how GN_30 performs exactly like Newton's method in terms of iterations. Addittionaly

- We can see a significantly better time cost;

- The memory needed for storing $S(x)$ is reduced by definition.

## 4.2  CGLS_k

In *Section* 3.1.1 we have already explored the usage of an iterative solver, in our case *Conjugate Gradient*, for the computation of the step of Newton's method. An interesting variant of *Conjugate Gradient* is the so-called **CGLS**, defined by Hestenes and Stiefel in [19]. The usage of CGLS is limited to *normal equations*, i.e., linear systems of the form

$$A^T A x = A^T b \qquad (4.7)$$

and it is considered more stable than Conjugate Gradient thanks to the recursive computation of the residual, and the avoidance of explicit formation of $A^T A$. When using the Gauss-Newton algorithm, we see that the linear system we aim to solve is exactly in the form of normal equations, having $A = J$ and $b = R$: in fact, it is quite common to use CGLS when using Gauss-Newton [6]. One downside of our proposed algorithm GN_k is, in fact, the impossibility of using such a method for solving the step computation linear system: this is due to the presence of the term $S$ in the left-hand side of the equation (4.5). In this section, we propose a variant of CGLS, namely **CGLS_k**, that adapts the linear solver to a system of the form

$$(A^T A + S) x = A^T b, \qquad (4.8)$$

and exploits this peculiar structure. The algorithm definition and its error analysis are inspired by another variant of CGLS, called CGLSc [7]. In *Algorithm 6* we define our linear solver. We see that the main difference between *Algorithm* 6 and CGLS is the presence of the term $v_k$, which is updated recursively to take into account the term $Sx$. To analyze the behavior of our algorithm on a linear system solving problem, we also propose an analysis on the structured conditioning of the problem, which is more informative than just the conditioning of ATA+S. The conditioning of (4.8) is the sensitivity of the solution $x$ to perturbations on the data $A, S, b$. We give an explicit formula for the structured condition number for perturbations on all of $A$, $S$, and $b$. In the following, we define the condition number of a function, as in [41].

**Algorithm 6** CGLS_k

    Input: A, S, b, $x_0$
    Define $r_0 = b - Ax_0$, $p_1 = s_0 = A^T r_0$, $v_0 = -Sx$
    **for** $k = 1, 2, ...$ **do**
        $t_k = Ap_k$
        $w_k = Sp_k$
        $\alpha_k = \|s_{k-1}\|^2 / (\langle t_k, t_k \rangle + \langle p_k, w_k \rangle)$
        $x_k = x_{k-1} + \alpha_k p_k$
        $r_k = r_{k-1} - \alpha_k t_k$
        $v_k = v_{i-1} - \alpha_k w_k$
        $s_k = A^T r_k + v_k$
        $\beta_k = \|s_k\|^2 / \|s_{k-1}\|^2$
        $p_{k+1} = r_k + \beta_k p_k$
    **end for**

**Definition 4.2.0.1.** *Let $\mathcal{X}$ and $\mathcal{Y}$ be normed vector spaces. if $F$ is a continuously differentiable function $F : \mathcal{X} \to \mathcal{Y}$, $x \mapsto F(x)$, the absolute condition number of $F$ at $x$ is the scalar $\|F'(x)\| := \sup_{\|v\|_{\mathcal{X}}=1} \|F'(x)v\|_{\mathcal{Y}}$, where $F'(x)$ is the Fréchet derivative of $F$ at $x$. The relative condition number of $F$ at $x$ is $\frac{\|F'(x)\| \|x\|_{\mathcal{X}}}{\|F(x)\|_{\mathcal{Y}}}$*

We consider $F$ as the function that maps $A, S, b$ to the solution $x$ of (4.8)

$$F : \mathbb{R}^{m \times n} \times \mathbb{R}^{n \times n} \times \mathbb{R}^m \to \mathbb{R}^n$$
$$(A, S, b) \mapsto F(A, S, b) = (A^T A + S)^{-1} b. \tag{4.9}$$

Given this, we can derive the following theorem, whose proof can be found in *Appendix* E.

**Theorem 4.2.1.** *The absolute condition number of the problem (4.8), with Euclidean norm on the solution and Frobenius norm (parametrized by $\alpha, \beta, \gamma$) on the data, is $\sqrt{\|\bar{M}\|}$, with $\bar{M} \in \mathbb{R}^{n \times n}$, given by*

$$\bar{M} = \left(\frac{\|x\|^2}{\gamma^2} + \frac{\|r\|^2}{\alpha^2}\right)(B^{-1}B^{-T}) + \left(\frac{1}{\beta^2} + \frac{\|x\|^2}{\alpha^2}\right)(B^{-1}A^T A B^{-T}) - \frac{1}{\alpha^2}(\tilde{B} + \tilde{B}^T), \tag{4.10}$$

*with $B = A^T A + S$, $\tilde{B} = B^{-1} A^T r x^T B^{-T}$ and $x$ the exact solution of (4.8).*

We use the parameters $\alpha, \beta, \gamma$ to indicate which perturbations we are taking into account, respectively on $A, S$, and $b$. If we assume that $(\alpha, \beta, \gamma) = (1, 1, 1)$, we define the structured relative condition number as

$$\kappa_S = \frac{\sqrt{\|\bar{M}\|} \|A, S, b\|_F}{\|x\|}, \qquad (4.11)$$

where $\bar{M}$ is defined in *Theorem* 4.2.1. This condition number can vary differently from $\kappa(A)$, since it also depends on $S$ and $b$: we will see this practically in the next section, performing some experiments on linear systems.

### 4.2.1 Experiments

To assess the soundness of CGLS_k, and show if it performs better than classical Conjugate Gradient, we ran some experiments solving linear systems with the different iterative solvers. The form of the linear systems is that of (4.8). To create different linear systems and control their conditioning, we define beforehand the solution $x^*$ and two sets of singular values $\sigma \in \mathbb{R}^n$, $\sigma_S \in \mathbb{R}^n$, then working as follows:

1. We set the floating-point precision to np.float64 - i.e, *double precision*;

2. We generate the matrices $U \in \mathbb{R}^{m \times m}$ and $V \in \mathbb{R}^{n \times n}$ of orthonormal bases performing QR-decomposition [43] of two different random matrices;

3. We form a diagonal matrix - in the SVD sense - $\Sigma \in \mathbb{R}^{m \times n}$, filling it with singular values $\sigma$;

4. We form $\mathbb{R}^{m \times n} \ni A = U\Sigma V^T$;

5. We generate the matrix $V_S \in \mathbb{R}^{n \times n}$ performing QR-decomposition on a new random matrix;

6. We form a diagonal matrix $\Sigma_S \in \mathbb{R}^{n \times n}$, filling it with singular values $\sigma_S$;

7. We form $\mathbb{R}^{n \times n} \ni S = V_S \Sigma_S \Sigma_S^T V_S^T$;

8. We solve the linear system $A^T b = (A^T A + S) x^*$ with respect to $b \in \mathbb{R}^m$ using a direct method.

After completing these steps, we have all the components needed to define the linear system. However, the final step (solving the system directly to compute the vector $b$) can be highly sensitive to numerical errors, especially if the matrices $A$ and $S$ are ill-conditioned. To reduce such errors, we perform all the steps in double precision. This higher precision helps minimize round-off and arithmetic inaccuracies. Then, when we run the iterative solver in lower precision (specifically, np.float32, i.e., *single precision*), such error becomes negligible. With this algorithm, we generated four different problems, using the set of singular values summarized in *Table* 4.1. Every array $\sigma$ and $\sigma_S$ contains $N = 10$ elements: each of them is proportionate to its index $i$. The solution $x^* \in \mathbb{R}^{10}$ is chosen to be equal to $[11, .., 1]$ for every problem. To assess the soundness of our algorithm, these sets of singular values have been chosen to cover the following different cases:

- **Pb.1** and **Pb.2**: slightly ill-conditioned $A$, with $S$ having negligible effect on conditioning;

- **Pb.3**: slightly ill-conditioned $A$, with $S$ improving conditioning;

- **Pb.4**: well-conditioned $A$, ill-conditioned $S$.

|  | $\sigma$ | $\sigma_S$ |
|---|---|---|
| **Pb. 1** | $[2.5^{i+1}]$ | $[(i+1) \times 10^{-5}]$ |
| **Pb. 2** | $[10^{i/3-5}]$ | $[(i+1) \times 10^{-10}]$ |
| **Pb. 3** | $[10^{i/3-5}]$ | $[(i+1) \times 10^5]$ |
| **Pb. 4** | $[10^{1+\frac{i}{N-1}}]$ | $[10^{-4+\frac{9i}{N-1}}]$ |

Table 4.1: Singular values of linear systems

On this set of linear systems, we used three different iterative solvers:

1. **CG**, the classical Conjugate Gradient;

2. **CGLS_k**, as defined in *Algorithm* 6;

3. what we called **CG_smart**, which is a version of CG in which the full matrix $A^TA+S$ is never formed explicitly and the matrix vector product is computed sequentially as in *CGLS*, but there is no usage of recursion as in $CGLS\_k$.

In the chart of *Figure* 4.3, we show the convergence to the solution $x^*$ of the three methods on **Pb.1**. As we can see from the chart in 4.3, $CGLS\_k$ not

Figure 4.3: Convergence of $CG$, $CG\_smart$ and $CGLS\_k$ on Pb.1

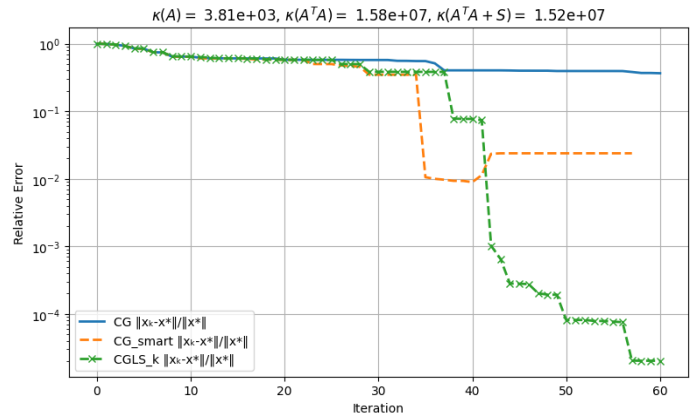

only has significantly better results than Conjugate Gradient on **Pb.1**, but also shows a good convergence speed. In the *Table* 4.2, we summarize the results on all the problems after 60 iterations, along with the conditioning numbers of the matrices and the structured condition number $\kappa_s$ of each system. We

| $\kappa(\mathbf{A})$ | $\kappa(\mathbf{A^TA+S})$ | $\kappa_\mathbf{s}$ | **CG** | **CG_smart** | **CGLS_k** |
|---|---|---|---|---|---|
| $4 \times 10^3$ | $2 \times 10^7$ | $1.6 \times 10^4$ | $3.7 \times 10^{-1}$ | $2.4 \times 10^{-2}$ | $\mathbf{2.0 \times 10^{-5}}$ |
| $1 \times 10^3$ | $1 \times 10^6$ | $4.5 \times 10^8$ | $6.6 \times 10^{-3}$ | $3.6 \times 10^{-3}$ | $\mathbf{6.7 \times 10^{-6}}$ |
| $1 \times 10^3$ | $1 \times 10^2$ | $1.8 \times 10^{24}$ | $\mathbf{6.5 \times 10^{-7}}$ | $1.5 \times 10^{-4}$ | $1.5 \times 10^{-4}$ |
| $1 \times 10^1$ | $4 \times 10^7$ | $2.7 \times 10^{13}$ | $9.3 \times 10^{-1}$ | $8.5 \times 10^{-1}$ | $\mathbf{8.4 \times 10^{-1}}$ |

Table 4.2: Results of iterative solvers

can see that $CGLS\_k$ shows a better performance than $CG$ and its variant

$CG\_smart$ on all problems, except for $Pb.3$, where the basic Conjugate Gradient is the best method. The specificity of the problem is that the full matrix $A^T A + S$ is better conditioned than the single matrices $A$ and $S$, and so it is more convenient to form the matrix to increase the stability of the iterative solving. This information is captured by our structured condition number $\kappa_s$: for the third problem, it shows a significantly higher value ($1.8 \times 10^{24}$) than for the others. We can also see from the table that our algorithm shows way better performance than the other methods in problems 1 and 2, where the matrix $A$ is slightly ill-conditioned, but forming the full matrix worsens its conditioning. A final remark is the fact that the structured condition number does not grow proportionally with relative error at the end of CGLS_k iteration: this is because we lack a backward error analysis as done in [7, Lemma 3.5]. Thanks to this estimate, we could make predictions on the forward error based *a priori*: this will be in our future work perspectives.

# Conclusion

The work presented focused on two research directions to improve the computational cost of Newton's method: developing an error analysis to derive a consistent Mixed-Precision Newton's algorithm, and proposing a new generalization of Gauss-Newton, presented as *GN_k*. The first contribution is the extension of the work of [45] to the minimization context, developing a framework that also covers *Quasi* and *Inexact* Newton approximations, deriving bounds on the problems' ill-conditioning - see (3.5) and (3.6) - and convergence rate (A.10) when using Newton's method with three different finite precisions. Thanks to the analysis, a bound on the stopping condition of the iterative linear solver used for step computation has been provided (3.17). Newton's and Gauss-Newton's algorithms have also been tested using mixed precisions on regression tasks, showing consistency with the predicted bounds on the accuracy, and also proving a better convergence on different datasets when compared to the *Adam* optimizer, with good performances on the test set. Additionally, the Gauss-Newton_k algorithm has been presented; it generalizes Newton's and Gauss-Newton's methods, and the theory on mixed-precision error analysis applies correctly to it. The minimizer allows for the computation of a subset of the second-order derivatives of the residual function in a Least-Squares context, converging also on non-linear problems, where Gauss-Newton fails, and reducing significantly the computational cost of classical Newton's method. An additional contribution is related to GN_k itself: to overcome the inability of using *CGLS* iterative linear solver, a new variant of Conjugate Gradient, called *CGLS_k*, has been presented. The solver has been tested on ill-conditioned linear problems, showing better performance than two other variants of Conjugate Gradient in the majority of cases.

## 4.3 Future work

The main open challenge regarding Mixed-Precision Newton's method - and its variants - is to apply it to real use-cases and problems. Even though it seems best fitted for small Machine Learning models, such as Logistic Regressors, its usage in the literature has also been extended to Neural Networks, especially for small nets, such as Physics-Informed ones [30, 24]. We also expect these problems to be a good fit for *GN_k*, as they tend to show significant imbalance of minimization hardness in the dimensions of their cost functions. For *Gauss-Newton_k*, we also would like to experiment with different ways of selecting which second-order derivatives to compute, for example, using a dynamic number of samples $k$, reducing as the algorithm converges. Another adaptive variant of Gauss-Newton that we started to explore focuses on using different precisions: instead of using the same finite precision for all $S_j$, we propose to use lower precisions for derivatives whose residual function is smaller, and vice versa. The approach has been tested on some toy problems with good results, also using low precisions such as bfloat16. Regarding $CGLS\_k$, an adaptation of the backward error analysis presented for *CGLSc* in [7] could be done, to predict the attainable accuracy of our solver and decide beforehand which iterative method to use for solving the Gauss-Newton_k step linear system. Another direction we find interesting is stochasticity; as we saw in *Section* 1, the datasets size is rapidly increasing over the years, and even though the usage of lower precisions mitigates the computational issue of training models on these data, it might still be necessary to use mini-batches during training. We tested the implementation of a stochastic Newton's method presented in [1] on the same regression tasks of *Section* 3.4 with remarkable results, also using the same low precisions used for standard Newton's method. Including a stochastic error analysis of such methods would allow us to better understand the potential and the limitations of the mixed-precision approach to batched second-order optimizers.

# Appendix A

# Theorem 3.1 - Proof

During the convergence proof, we will be using the following *Lemma*

**Lemma A.0.1.** *For any $v, w \in \mathbb{R}^n$,*

$$\|g(w) - g(v) - H(v)(w-v)\| \leq \frac{L_H}{2}\|w-v\|^2 \qquad (A.1)$$

The proof of the lemma can be found in [10, Lem. 4.1.12]

*Proof.* For simplicity, let's say

$$\nu := \phi(g, \hat{x}_i, n, u_l, u)\kappa(H(\hat{x}_i)),$$
$$\mu := L_H \|H_*^{-1}\| \|x - x^*\|.$$

We can then note, from definition of $E^H$ (3.3), that

$$\begin{aligned}
\|H^{-1}E^H\| &\leq \|H^{-1}\|\phi(g, \hat{x}_i, n, u_l, u)\|H\| \\
&= \phi(g, \hat{x}_i, n, u_l, u)\kappa(H(\hat{x}_i)) = \nu \\
&\leq \frac{1}{8}.
\end{aligned} \qquad (A.2)$$

We start from the Assumption 3.6 and Lipschitz continuity of Hessian (2.2)

$$\|H_*^{-1}(H - H_*)\| \leq L_H \|H_*^{-1}\| \|x - x^*\| = \mu \leq \frac{1}{8}. \qquad (A.3)$$

Then, using identity,

$$H = H_*(I + H_*^{-1}(H - H_*)), \tag{A.4}$$

thanks to this and the assumption that $H_*$ is nonsingular, we have that $H$ is nonsingular with inverse given by

$$H^{-1} = (I + H_*^{-1}(H - H_*))^{-1} H_*^{-1}.$$

Using norm properties on this definition and A.3, we have

$$\|H^{-1}\| \leq \frac{\|H_*^{-1}\|}{1 - \|H_*^{-1}(H - H_*)\|} \leq \frac{1}{1 - \mu}\|H_*^{-1}\|. \tag{A.5}$$

We can do a similar procedure for $H + E^H$, then using both A.5 and A.2

$$\|(H + E^H)^{-1}\| \leq \frac{\|H^{-1}\|}{1 - \|H^{-1}E^H\|} \leq \frac{1}{(1 - \mu)(1 - \nu)}\|H_*^{-1}\|. \tag{A.6}$$

This leads to a stricter result than the one presented in [28, Lem. 2.2]: in fact, the latter lacks the term $(1 - \mu)$. Granting the nonsingularity of $H + E^H$, we are also proving the well-definedness of $\bar{x}$. We have

$$\begin{aligned}\bar{x} - x^* &= x - x^* - (H + E^H)^{-1}(g + E^g) + E^+ \\ &= (I - (H + E^H)^{-1}H)(x - x^*) \\ &\quad - (H + E^H)^{-1}(g - H(x - x^*) + E^g) + E^+,\end{aligned}$$

Then putting in norms

$$\begin{aligned}\|\bar{x} - x^*\| &\leq \|I - (H + E^H)^{-1}H\|\|x - x^*\| \\ &\quad - \|(H + E^H)^{-1}\|(\|g - H(x - x^*)\| + \|E^g\|) + \|E^+\|.\end{aligned} \tag{A.7}$$

We now reformulate the first term of the right-hand side as follows

$$I - (H + E^H)^{-1}H$$
$$= (H + E^H)^{-1}(H + E^H) - (H + E^H)^{-1}H$$
$$= (H + E^H)^{-1}E^H$$
$$= (I + H^{-1}E^H)^{-1}H^{-1}E,$$

since $H + E^H = H(I + H^{-1}E^H)$. From this reformulation and A.2, it follows that

$$\|I - (H + E^H)^{-1}H\| \leq \frac{1}{1 - \|H^{-1}E^H\|}\|H^{-1}E^H\| = \frac{1}{1 - \nu}\|H^{-1}E^H\|.$$

For the second term of A.7, we use *Lemma* A.0.1: thanks to it we have

$$\|g - H(x - x^*)\| \leq \frac{L_H}{2}\|x - x^*\|^2 \text{ and}$$
$$\|g - H_*(x - x^*)\| \leq \frac{L_H}{2}\|x - x^*\|^2,$$

And we use these disequations to bound the norm of $g$

$$\|g\| \leq \|g - H_*(x - x^*)\| + \|H_*(x - x^*)\|$$
$$\leq \frac{L_H}{2}\|x - x^*\|^2 + \|H_*\|\|x - x^*\|,$$

and hence, from 3.2,

$$\|E^g\| \leq u(\frac{L_H}{2}\|x - x^*\|^2 + \|H_*\|\|x - x^*\|) + \psi(g, x, u, u_g). \qquad (A.8)$$

From the bound on $E^+$ (3.4) we have

$$\|E^+\| \leq u(\|x - x^*\| + \|x^*\| + \|d\|),$$

with

$$\begin{aligned}
\|d\| &\leq \|(H+E^H)^{-1}\|(\|g\|+\|E^g\|) \\
&\leq \|(H+E^H)^{-1}\|((1+u)\|g\|+\psi(g,x,u,u_g)) \\
&\overset{\text{Using A.6 and A.8}}{\leq} \frac{1}{(1-\mu)(1-\nu)}\|H_*^{-1}\| \\
&\quad \left[(1+u)\left(\frac{L_H}{2}\|x-x^*\|+\|H_*\|\right)\|x-x^*\|+\psi(g,x,u,u_g)\right].
\end{aligned} \quad (A.9)$$

So finally, starting from A.7 and using the derived bound, we have that

$$\|\bar{x}-x^*\| \leq G\|x-x^*\| + lim_{acc}, \quad (A.10)$$

where

$$G = \frac{1}{1-v}\|H^{-1}E^H\| + \frac{(1+u)^2}{2(1-\mu)(1-\nu)}L_H\|H_*^{-1}\|\|x-x^*\|+$$
$$+\frac{u(2+u)}{(1-\mu)(1-\nu)}\kappa(H_*) + u$$

and

$$lim_{acc} = \frac{1+u}{(1-\mu)(1-\nu)}\|H_*^{-1}\|\psi(g,x,u,u_g) + u\|x^*\|$$

We also stated that $\mu \leq \frac{1}{8}$ and $\nu \leq \frac{1}{8}$ in *Assumptions* 3.5,3.6, so it is notable that $G < 1$. In this case, the error contracts linearly until $lim_{acc} \gtrsim \|x-x^*\|$, when the latter term of the disequation makes the former one negligible. Hence, the relative error can be guaranteed to decrease until it reaches $lim_{acc}/\|x^*\|$. □

We can see that assumptions 3.5 and 3.6 are necessary to prove the existence of $\bar{x}$, and that assumption 3.5 is a bound on the error made in the formation of the hessian and the sability of the linear solver, other than on the ill-conditioning of the problem. Using exact Newton and exact arithmetic, $u = \psi(g,x,u,u_g) = \nu = 0$ and $E^H = 0$. Then, for $\mu \leq 1/2$, we find again the local quadratic convergence of Newton, having the same assumption as the proof in [36].

Relating our result to that of [28, Eq. 2.13], if we consider using double, single, and half precisions respectively as $\pi_g, \pi_w, \pi_l$, we can see they are pretty similar. As also cited in the article, the relative error has a term independent of the error in the previous step, but is related to the approximation of the gradient.

# Appendix B

# Theorem 3.2 - Proof

*Proof.* For simplicity, let's say

$$\nu := \phi(g, \hat{x}_i, n, u_l, u)\kappa(H(\hat{x}_i)),$$

$$\mu := L_H \|H_*^{-1}\| \|x - x^*\|.$$

Thanks to 3.8 and 3.10 we can use *Theorem* 3.1.1 to state that $\bar{x}$ is well defined. Let $\bar{g} = g(\bar{x})$ and define $w \in \mathbb{R}^n$ by $w = \bar{g} - g - H(\bar{x} - x)$. Note that $\bar{x} - x = -d + E^+$ and $Hd = g + E^g - E^H d$ from Newton's step equation 3.1. This leads to

$$\bar{g} = g + H(-d + E^+) + w = -E^g + E^H d + HE^+ + w,$$

which yields

$$\begin{aligned}\|\bar{g}\| &\leq \|E^g\| + \|E^H\|\|d\| + \|H\|\|E^+\| + \|w\| \\ &\leq u\|g\| + \psi(g, x, u, u_g) + \|H\|\|d\|(\phi(g, x, n, u_l, u) + u) \quad \text{(B.1)} \\ &\quad + u\|H\|\|x\| + \|w\|.\end{aligned}$$

Reusing A.3 and A.4, we have that

$$\|H\| \leq (1 + \mu)\|H_*\|. \quad \text{(B.2)}$$

Using A.9, A.6 and A.2 we have

$$\|d\| \leq \|(H + E^H)^{-1}\|(\|g\| + \|E^g\|) \leq \frac{1}{1-\nu}\|H^{-1}\|((1+u)\|g\| \quad \text{(B.3)}$$
$$+ \psi(g, v, u, u_g)),$$

which gives, using A.5 and B.2, and starting from the third term of B.1,

$$\|H\|\|d\|(\phi(g, x, n, u_l, u) + u) \leq \frac{(1+\mu)(1+u)}{(1-\mu)(1-\nu)}\kappa(H_*)\{\phi(g, x, n, u_l, u) + u\}\|g\|$$
$$+ \frac{1+\mu}{(1-\mu)(1-\nu)}\kappa(H_*)\{\phi(g, x, n, u_l, u) + u\}\psi(g, x, u, u_g).$$
$$\text{(B.4)}$$

From *Lemma* A.0.1 we have

$$\|w\| \leq \frac{L_H}{2}\|\bar{x} - x\|^2. \quad \text{(B.5)}$$

First, we know from the definition of Newton step with errors (3.1)

$$\bar{x} = x + d + \Delta x + \Delta d \implies$$
$$\|\bar{x} - x\| \leq \|d\| + u\|x\| + u\|d\|,$$

starting from this, and using A.5 and B.3,

$$\|\bar{x} - x\| \leq (1+u)\|d\| + u\|x\|$$
$$\leq \|H_*^{-1}\| \left(\frac{(1+u)^2}{(1-\mu)(1-\nu)}\|g\| + \frac{1+u}{(1-\mu)(1-\nu)}\psi(g, x, u, u_g)\right)$$
$$+ u\|x\|.$$
$$\text{(B.6)}$$

Starting from triangle inequality and *Theorem* 3.1.1 we have

$$\|\bar{x} - x\| \leq (G+1)\|x - x^*\| + lim_{acc}. \quad \text{(B.7)}$$

We can then multiply the results from B.6 and B.7 to have the needed $\|\bar{x} - x\|^2$ of B.5, as follows:

$$\begin{aligned}
\|w\| \leq &\frac{(1+u)^2(G+1)}{2(1-\mu)(1-\nu)} L_H \|H_*^{-1}\| \|x - x^*\| \|g\| \\
&+ \frac{(1+u)^2}{2(1-\mu)(1-\nu)} L_H \|H_*^{-1}\| lim_{acc} \|g\| \\
&+ \frac{(1+u)(G+1)}{2(1-\mu)(1-\nu)} L_H \|H_*^{-1}\| \|x - x^*\| \psi(g, x, u, u_g) \\
&+ \frac{(1+u)}{2(1-\mu)(1-\nu)} L_H \|H_*^{-1}\| lim_{acc} \psi(g, x, u, u_g) \\
&+ \frac{G+1}{2(1-\mu)} L_H \|H_*^{-1}\| \|x - x^*\| u \|H\| \|x\| \\
&+ \frac{1}{2(1-\mu)} L_H lim_{acc} \|H_*^{-1}\| u \|H\| \|x\|,
\end{aligned} \quad (\text{B.8})$$

where the penultimate and the last terms on the right-hand side of the inequality are obtained using $\|H\| \|H^{-1}\| \geq 1$ and A.5. Substituting B.4 and B.8 into B.1 we have

$$\|\bar{g}\| \leq P \|g\| + lim_g,$$

with $P$ and $lim_g$ defined as follows

$$P = c_0 \left[ \mu + \tau + u\kappa(H_*) \right],$$

and

$$lim_g = c_1(\mu + \tau + u\kappa(H_*))\psi(g, x, u, u_g) + c_2(\mu + \tau + 1)u\|H\|\|x\|,$$

where

$$\tau = L_H lim_{acc} \|H_*^{-1}\|,$$

with $lim_{acc}$ as defined in *Theorem* 3.1.1 and $c_0$, $c_1$ and $c_2$ constants of order 1. Thanks to *Assumptions* 3.8, 3.9 and 3.10 we have that $P < 1$. This means that the gradient norm decreases linearly until $p \gtrsim \|g\|$, where the $P$ term becomes negligible. At this point, still using the theorem assumptions, we see that $p \approx \psi(g, \hat{x}_i, u, u_g) + u\|H(\hat{x}_i)\|\|\hat{x}_i\|$: thus, as stated in the theorem definition, this is the granted limit to gradient norm. □

# Appendix C

# Inexact Newton model equivalence

In this section, it will be numerically proven that both the Inexact Newton model and the Hessian backward model can generalize, and vice versa: this latter proof will be beneficial, because it allows it to include iterative solving error in our $E^H$ upper-bound $\phi$.

First, we see how the Inexact Newton model generalizes the backward error model on $H$ in our error analysis. We begin by writing down the Newton step linear system for minimization: note that we formalize the error in the system solving as a backward error on $H$ (see 3.3). We will consider the standard *residual* $r$ definition, as follows,

$$\|r\| = \|H\hat{d} + g\| \leq \eta\|g\|. \tag{C.1}$$

We avoid the explicit writing of $\hat{x}$, thus $H = H(\hat{x})$ and $g = g(\hat{x})$. Also, we denote $\phi(g, \hat{x}, n, u_l, u)$ simply as $\phi$. First, we define a value $\sigma$ as

$$\sigma := \frac{\|H\hat{d}\|}{\|H\|\|\hat{d}\|} = \frac{\|H\hat{d}\|}{\|H\|\|H^{-1}H\hat{d}\|} \geq \frac{\|H\hat{d}\|}{\kappa(H)\|H\hat{d}\|} = \frac{1}{\kappa(H)}, \tag{C.2}$$

Thus, $\sigma \leq 1$ due to the range of values of $\kappa(H)$. By definition of $r$, we have that

$$H\hat{d} + g = r, \tag{C.3}$$

and we can write Newton's step linear system as

$$(H + E^H)\hat{d} = -g \implies H\hat{d} + E^H\hat{d} = -g. \tag{C.4}$$

By subtracting C.3 from C.4, we have

$$E^H\hat{d} = -r \implies \|E^H\hat{d}\| = \|r\|. \tag{C.5}$$

Using the definition of $E^H$ (see 3.3), we can derive

$$\begin{aligned} \|r\| = \|E^H\hat{d}\| &\leq \phi\|H\|\|\hat{d}\| \\ &\leq \phi\|g\|\frac{\|H\|\|\hat{d}\|}{\|g\|} \\ &= \phi\|g\|\underbrace{\frac{\|H\|\|d\|}{\|H\hat{d}\|}}_{1/\sigma}, \end{aligned} \tag{C.6}$$

so, $\|r\| \leq \phi\|g\|\frac{1}{\sigma}$. If we define the iterative solver stopping condition $\eta = \phi\frac{1}{\sigma}$, we can see how, starting from our model, we reduce the backward error on $H$ as part of the *Inexact Newton* framework.

We now prove that our definition of backward error on $H$ generalizes *Inexact Newton* error, showing an equivalence of the two models. In other words, we now want to derive a bound on backward error $E^H$, using *Inexact Newton* model and generalize it by the usage of our bound $\phi$. To do so, we use the *Rigal-Gaches* backward-error definition (See [20, Ch. 7], as follows:

$$\Delta A = \frac{(b - A\hat{x})x^T}{\|x\|_2^2}.$$

In our frame, we have:

$$\text{(I)} \ A := H,$$

$$\text{(II)} \ \Delta A := E^H,$$

$$\text{(III)} \ \hat{x} := \hat{d},$$

$$\text{(IV)} \ b := -g,$$

and we remind

$$\begin{aligned}(H + E^H)\hat{d} &= -g, \\ H\hat{d} + g &= r.\end{aligned} \tag{C.7}$$

Then, starting with backward error bounding,

$$\begin{aligned}\|E^H\| &\leq \frac{\|r\|\|\hat{d}\|}{\|\hat{d}\|^2} = \frac{\|r\|}{\|\hat{d}\|} \\ &\stackrel{\text{using 2.4}}{\leq} \eta \frac{\|g\|}{\|\hat{d}\|} = \eta \frac{\|H\hat{d}\|}{\|\hat{d}\|} \\ &= \eta \|H\| \underbrace{\frac{\|H\hat{d}\|}{\|H\|\|\hat{d}\|}}_{\sigma},\end{aligned} \tag{C.8}$$

which means $\|E^H\| \leq \eta\sigma\|H\|$. Knowing that $\|E^H\| \leq \phi\|H\|$, we can define $\phi = \eta\sigma$ - as we have also done previously - and see how our model 3.3 can generalize the *Inexact Newton* bounding. We can also see how formalizing the iterative solving of the Newton linear system as a backward error $H$, including it in term $\phi$, excludes its inexactness from limiting the accuracy of *Theorems* 3.1.1, 3.1.2. However, this term still affects the convergence rate of our mixed-precision method.

# Appendix D

# Gauss-Newton proofs

Firstly, we want to prove

$$(J(x) - J(x^*))^T R(x^*) \cong S(x^*)(x - x^*), \tag{D.1}$$

which can be demonstrated as follows. Let's start by noting that

$$\begin{aligned}(J(x) - J(x^*))^T R(x^*) &= J(x)^T R(x^*) - \underbrace{J(x^*)^T R(x^*)}_{=0} \\ &= J(x)^T R(x^*) \\ &= \sum_{i=1}^{m} r_i(x^*) \nabla r_i(x),\end{aligned} \tag{D.2}$$

and rewrite $S(\theta^*)$ as it is defined in 3.22, then, substituting in D.1,

$$\begin{aligned}\sum_{i=1}^{m} \cancel{r_i(x^*)} \nabla r_i(x) &= \sum_{i=1}^{m} \cancel{r_i(x^*)} \nabla^2 r_i(x^*)(x - x^*) \\ \implies \sum_{i=1}^{m} \nabla r_i(x) &= \sum_{i=1}^{m} \nabla^2 r_i(x^*)(x - x^*).\end{aligned} \tag{D.3}$$

We can write the first-order Taylor expansion of the gradient $\nabla r(x)$, assuming $\lim x \to x^*$, as follows,

$$\nabla r(x) = \nabla r(x^*) + \nabla^2 r(x^*)(x - x^*) + O(\|x - x^*\|^2)$$
$$\underset{\nabla r(x^*) \approx 0}{\Longrightarrow} \nabla r(x) \approx \nabla^2 r(x^*)(x - x^*) \tag{D.4}$$

We can see that the two terms are approximately equal by using the Taylor expansion of $\nabla r(x)$ at $x^*$, and summing the components, with the assumptions that

1. $\nabla r(x^*) = 0$, which is approximately true, as $x^*$ is the solution of our minimization problem;

2. $x \to x^*$, which means that we are close enough to the solution.

From this, we can also rewrite the condition for the Gauss-Newton convergence as

$$\|S(x^*)(x - x^*)\| \leq \sigma \|x - x^*\|, \quad \forall \theta \in D. \tag{D.5}$$

The convergence theorem reported in [10] for the Gauss-Newton method also requires assumptions regarding $J$ Lipschitz-continuity, bounded by a constant $\gamma$, and its upper-bound $\alpha$: these are useful to state Lipschitz-continuity of $J^T J$, which is helpful in the following proofs about the convergence of mixed-precision Gauss-Newton. We can see that,

$$\begin{aligned}
\|J(x)^T J(x) - J(y)^T J(y)\| &\leq \|J(x)^T (J(x) - J(y)) + (J(x)^T - J(y)^T) J(y)\| \\
&\leq \|J(x)^T (J(x) - J(y))\| + \|(J(x)^T - J(y)^T) J(y)\| \\
&\leq \|J(x)^T\| \|J(x) - J(y)\| + \|J(x)^T - J(y)^T\| \|J(y)\| \\
&\leq 2\alpha\gamma \|x - y\|,
\end{aligned} \tag{D.6}$$

thus, $J(x)^T J(x)$ is Lipschitz continuous with Lipschitz constant $L = 2\alpha\gamma$, where $\alpha$ is an upper-bound on norm of $J(x)$ and $\gamma$ is the Lipschitz-constant of $J(x)$.

To prove that the assumptions on convergence of Gauss-Newton imply our *Theorem* 3.1.1 assumptions, we can first explicitly include the approximation error on the Hessian in the term $\phi$. Assuming to solve the step computation linear system directly, we can write

$$\phi(g, x_i, n, u_l, u) \approx u_l + \sigma \|H\|^{-1}, \tag{D.7}$$

where $u_l$ is the error coming from finite precision, and $\sigma$ is the absolute value upper-bounding the discarded term $S(x)$. During the proof, definitions from [10, Th. 10.2.1] will be used, in particular for $c$ and $\sigma$ terms.

*Proof.* First, in *Eq.* D.6, we have already shown that $H(x) = J(x)^T J(x)$ is $L_H$-continuous, where $L_H \leq 2\alpha\gamma$, and $\alpha$ and $\gamma$ are an upper-bound on $\|J(x)\|$ and the Lipschitz constant of $J(x)$, respectively. We can also see $g(x^*) = J(x^*)^T R(x^*) = 0$ by definition of $g(x)$ and $x^*$. We now prove the first two inequalities. For 3.5, we can use the definition of $\phi$ D.7, as follows,

$$\begin{aligned}\phi\kappa(H) &\leq \sigma \|H\|^{-1} \|H\| \|H^{-1}\| \\ &\leq \sigma \|H^{-1}\| \\ &\leq \sigma \cdot \frac{c}{\lambda} \\ &< \sigma \cdot \frac{\lambda}{\sigma} \cdot \frac{1}{\lambda} = 1,\end{aligned} \tag{D.8}$$

where, for the last inequality, we use the definition of $c$, and for the second inequality we use [10, page 10.2.7]

For *Ass.* 3.6, we have that,

$$L_H \|H_*^{-1}\| \|x - x^*\| \leq 2\alpha\gamma \|H_*^{-1}\| \frac{\lambda - c\sigma}{c\alpha\gamma}, \tag{D.9}$$

by D.6 and [10, Eq. 10.2.8]. We also know that $\|A^{-1}\|_2 \leq \frac{1}{\lambda}$, where $\lambda$ is the minimum eigenvalue of $A$. In our case, we know that the smallest eigenvalue of $H(x^*) = J(x^*)^T J(x^*)$ is $\lambda$, thus,

$$
\begin{aligned}
L_H \|H_*^{-1}\| \|x - x^*\| &\leq 2c\alpha \frac{1}{\lambda} \frac{\lambda - c\sigma}{c\alpha} \\
&= 2 \cdot \frac{1}{\lambda} \cdot \left(\frac{\lambda}{c} - \frac{\cancel{c}\sigma}{\cancel{c}}\right) \\
&= 2 \left(\frac{1}{c} - \frac{\sigma}{\lambda}\right).
\end{aligned} \tag{D.10}
$$

So we need to prove that $\frac{1}{c} - \frac{\sigma}{\lambda} \leq 2$. Since $c$ is a free parameter in the range $(1, \lambda/\sigma)$, we can freely choose a $c$ s.t. $\lambda/\sigma < 1/c \leq 3\sigma/2\lambda$ to prove the assumption, as follows,

$$
\begin{aligned}
L_H \|H_*^{-1}\| \|x - x^*\| &\leq 2 \left(\frac{1}{c} - \frac{\sigma}{\lambda}\right) \\
&\leq 2 \left(\frac{3}{2} \cdot \frac{\sigma}{\lambda} - \frac{\sigma}{\lambda}\right) \\
&= \frac{\sigma}{\lambda} < 1.
\end{aligned} \tag{D.11}
$$

$\square$

# Appendix E

# Structured conditioning for CGLS_k

Let's start with the definition of the structured condition number of a function:

**Definition E.0.0.1.** *Let $\mathcal{X}$ and $\mathcal{Y}$ be normed vector spaces. If $F$ is a continuously differentiable function*

$$F : \mathcal{X} \to \mathcal{Y}, \quad x \mapsto F(x), \tag{E.1}$$

*the absolute condition number of $F$ at $x$ is the scalar*

$$\|F'(x)\| := \sup_{\|v\|_{\mathcal{X}}=1} \|F'(x)v\|_{\mathcal{Y}}, \tag{E.2}$$

*where $F'(x)$ is the Fréchet derivative of $F$ at $x$. The relative condition number of $F$ at $x$ is*

$$\frac{\|F'(x)\|\|x\|_{\mathcal{X}}}{\|F(x)\|_{\mathcal{Y}}} \tag{E.3}$$

We consider $F$ as the function that maps $A, S, b$ to the solution $x$ of our linear system

$$(A^T A + S)x = A^T b, \tag{E.4}$$

hence,
$$(A, S, b) \mapsto F(A, S, b) = (A^T A + S)^{-1} A^T b. \tag{E.5}$$

The Fréchet derivative in finite-dimensional spaces is the usual derivative. In particular, we represent it with the Jacobian matrix. If $F$ is Fréchet differentiable at a point $(A, S, b)$, the its derivative is

$$\begin{aligned} F'(A, S, b) &: \mathbb{R}^{m \times n} \times \mathbb{R}^{n \times n} \times \mathbb{R}^m \\ F'(A, S, b)(E, E_S, g) &= J_F(A, S, b)(E, E_S, g), \end{aligned} \tag{E.6}$$

where $J_F(A, S, b)(E, E_S, g)$ denotes the Jacobian matrix of $F$ at $(A, S, b)$ applied to $(E, E_S, g)$. As by *Def.* E.0.0.1, the condition number of $F$ at the point $(A, S, b)$ is given by

$$\|F'(A, S, b)\| = \sup_{\|(E, E_S, g)\|_{F(\alpha, \beta, \gamma)} = 1} \|F'(A, S, b)(E, E_S, g)\|, \\ E \in \mathbb{R}^{m \times n}, E_S \in \mathbb{R}^{n \times n}, g \in \mathbb{R}^m. \tag{E.7}$$

Using parametrized Frobenius norm we can easily decide what to perturbe thanks to parameters $\alpha, \beta, \gamma$. For instance, we can decide to perturbe $A$ and $b$ only choosing a large $\gamma$. In fact, the condition $\gamma \to \infty$ implies $E_S \to 0$, from the constraint $\alpha^2 \|E\|_F^2 + \gamma^2 \|E_S\|_F^2 + \beta^2 \|g\|_F^2 = 1$ in the definition of the condition number.

We can say $A$ to be perturbed to $\tilde{A} = A + E$, $S$ to $\tilde{S} = S + E_S$, and $b$ to $\tilde{b} = b + g$. Then, $B = (A^T A + S)$ is perturbed to

$$\tilde{B} = \tilde{A}^T \tilde{A} + \tilde{S} = (A+E)^T(A+E) + (S+E_S) = A^T A + A^T E + E^T A + S + E_S \tag{E.8}$$

neglecting second-order terms ($E^2$). The soluton $x = (A^T A + S)^{-1} A^T b$ is then perturbed to $\tilde{x} = x + \delta x = (\tilde{A}^T \tilde{A} + \tilde{S})^{-1} \tilde{A}^T \tilde{b}$. Then, $\tilde{x}$ solves

$$(A^T A + A^T E + E^T A + S + E_S)\tilde{x} = (A^T b + E^T b + A^T g). \tag{E.9}$$

Knowing that $r = b - Ax$, an neglecting second-order terms We can reformulate this equation as follows

$$E^T A(x + \delta x) - E^T b + (A^T A + A^T E + S + E_S)(x + \delta x) = A^T b + A^T g$$
$$\implies -E^T r + \underbrace{Bx - A^T b}_{=0} + B\delta x + A^T Ex + E_S x = A^T g$$
$$\implies B\delta x = E^T r - A^T Ex - E_S x + A^T g$$
$$\implies \delta x = B^{-1} E^T r + B^{-1} A^T (g - Ex) - B^{-1} E_S x. \tag{E.10}$$

We then conclude that

$$F'(A, S, b)(E, E_S, g) = B^{-1} E^T r + B^{-1} A^T (g - Ex) - B^{-1} E_S x, \tag{E.11}$$

for all $E \in \mathbb{R}^{m \times n}$, $E_S \in \mathbb{R}^{n \times n}$ and $g \in \mathbb{R}^m$. We can then prove *Theorem* 4.2.1, that gives an explicit and computable formula for the structured condition number.

*Proof.* Let us consider

$$y^T F'(A, S, b)(E, E_S, g) = y^T B^{-1} E^T r - y^T B^{-1} A^T Ex + y^T B^{-1} A^T g - y^T B^{-1} E_S x$$
$$= r^T E B^{-1} y - y^T B^{-1} A^T Ex + y^T B^{-1} A^T g - x^T E_S^T B^{-1} y. \tag{E.12}$$

We recall that $\mathbb{R}^{m \times n} \ni E = \sum_{i=1}^{n} \sum_{j=1}^{m} e_i^T E e_j \Delta_{ij}$, where $e_i, e_j$ are the $i$-th and $j$-th vectors of the canonical basis, and $\mathbb{R}^{m \times n} \ni \Delta_{ij} = e_i e_j^T$.

Then, we can rewrite the expression as

$$\left[ \frac{\text{vec}(R)}{\alpha}, \frac{y^T B^{-1} A^T}{\beta}, \frac{\text{vec}(\tilde{R})}{\gamma} \right] \cdot [\alpha \text{vec}(E), \beta g, \gamma \text{vec}(E_S)]^T := \\ w^T [\alpha \text{vec}(E), \beta g, \gamma \text{vec}(E_S)]^T, \tag{E.13}$$

introducing the matrices $R$ and $\tilde{R}$, such that

$$R_{ij} = r^T \Delta_{ij} B^{-1} y - y^T B^{-1} A^T \Delta_{ij} x,$$
$$\tilde{R}_{ij} = x^T \Delta ij B^{-1} y \tag{E.14}$$

Given this, we know that $w = \|F'(A, S, b)\| y$, so we focus on its norm. The squared norm of $\text{vec}(R)$ follows as

$$\begin{aligned}
\|\text{vec}(R)\|^2 &= \sum_{i=1}^{n} \sum_{j=1}^{m} (r^T \Delta_{ij} B^{-1} y - y^T B^{-1} A^T \Delta_{ij} x)^2 \\
&= \|r\|^2 \|B^{-1} y\|^2 + \|x\|^2 \|AB^{-T} y\|^2 \\
&\quad - y^T (B^{-1} A^T r x^T B^{-T} + B^{-1} x r^T A B^{-T}) y,
\end{aligned} \tag{E.15}$$

and the one of $\text{vec}(\tilde{R})$ is equal to

$$\begin{aligned}
\|\text{vec}(\tilde{R})\|^2 &= \sum_{i=1}^{n} \sum_{j=1}^{m} (x^T \Delta_{ij} B^{-1} y)^2 \\
&= \|x\|^2 \|B^{-1} y\|^2.
\end{aligned} \tag{E.16}$$

Putting all together, we have that

$$\begin{aligned}
\|w\|^2 &= y^T \bar{M} y, \\
\bar{M} &:= \left(\frac{\|x\|^2}{\gamma^2} + \frac{\|r\|^2}{\alpha^2}\right) B^{-2} + \left(\frac{1}{\beta^2} + \frac{\|x\|^2}{\alpha^2}\right) (B^{-1} A^T A B^{-T}) - \frac{1}{\alpha^2} (\tilde{B} + \tilde{B}^T) \\
\tilde{B} &:= B^{-1} A^T r x^T B^{-T}.
\end{aligned} \tag{E.17}$$

Thus, we finally have that

$$\|F'(A, S, b)\| = \sqrt{\|\bar{M}\|}. \tag{E.18}$$

. □